\documentclass[conference]{IEEEtran}
\usepackage{times}
\usepackage[numbers]{natbib}
\usepackage{multicol}
\usepackage[utf8]{inputenc}
\usepackage[T1]{fontenc}
\usepackage{amsmath,amssymb}
\usepackage{graphicx}
\usepackage{capt-of}   %
\usepackage{booktabs}
\usepackage{xcolor}
\usepackage{siunitx}
\usepackage{tikz}
\usetikzlibrary{arrows.meta,positioning}
\usepackage[bookmarks=true,hidelinks]{hyperref}
\usepackage[capitalise]{cleveref}

\newcommand{\method}{E2S2H}

\hypersetup{
  pdftitle={\method: Agentic Real-to-Sim for Learning Loco-Manipulation VLAs without Robot Demonstrations},
  pdfauthor={Chen Liu, Nikolaus Correll}}

\begin{document}

\title{\includegraphics[height=1.0in,trim=51 124 44 158,clip]{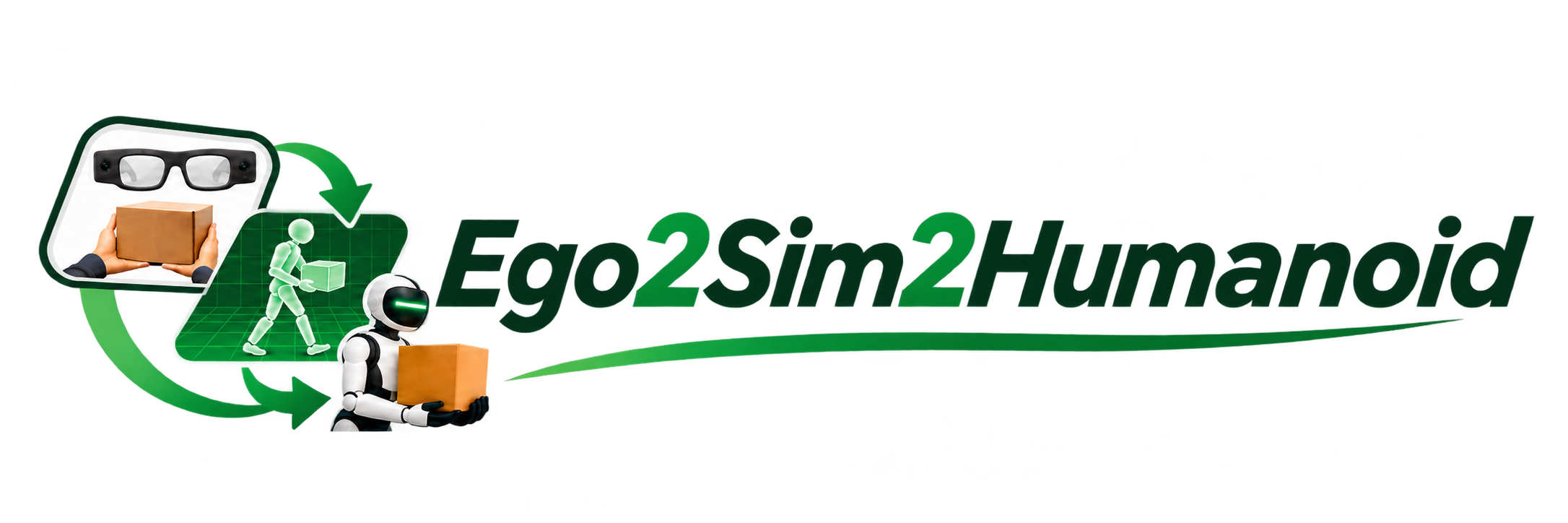}\\[3pt]
\LARGE Agentic Real-to-Sim\\for Learning Loco-Manipulation VLAs\\without Robot Demonstrations}

\author{\authorblockN{Chen Liu and Nikolaus Correll}
\authorblockA{University of Notre Dame}}

\makeatletter
\newif\if@teaserset
\let\paper@maketitle\@maketitle
\def\@maketitle{\paper@maketitle\paper@teaser}
\def\paper@teaser{%
  \if@teaserset\addtocounter{figure}{-1}\fi
  \vspace{-0.4\baselineskip}%
  \begin{center}
    \includegraphics[width=\textwidth]{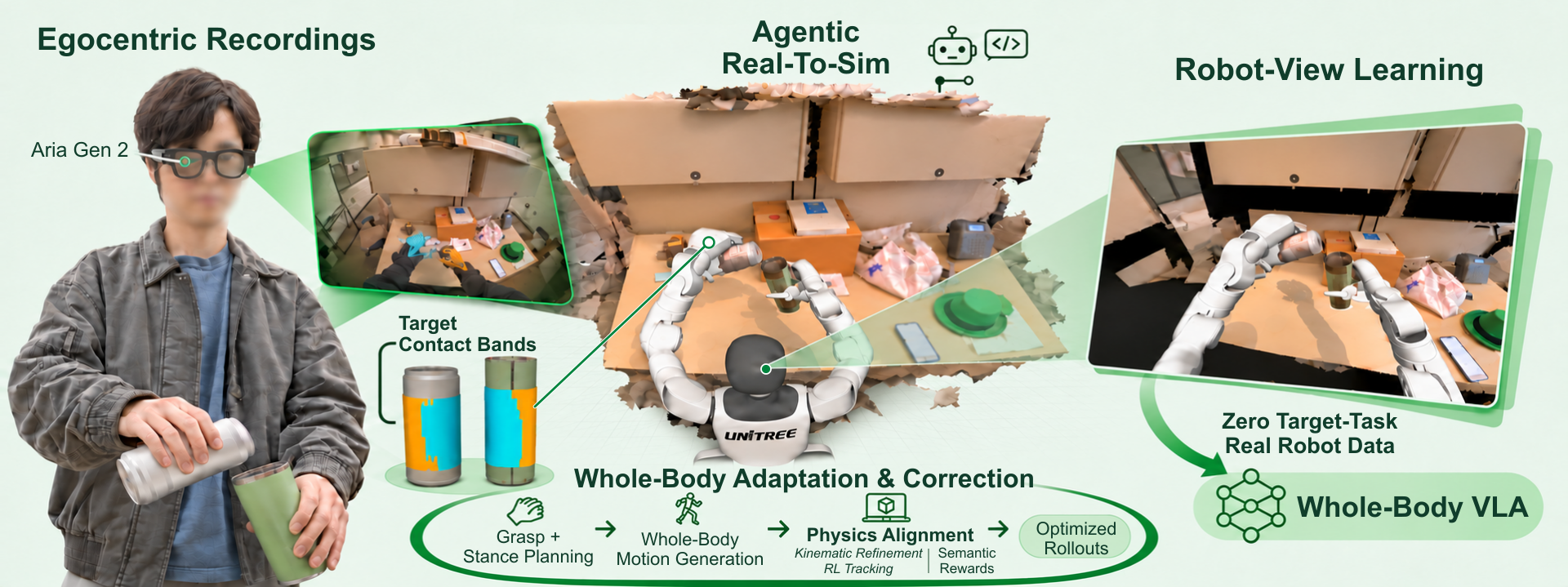}
    \captionof{figure}{Overview of \method{}. Egocentric glasses supply the head
    trajectory, in-view hand poses and RGB-D streams in one metric frame.
    Agentic real-to-sim reconstructs each recording without a separate pre-scan or CAD templates, building collision models and textured visual meshes for the scene and tracked objects, completing unseen surfaces when
    needed, and localizing target contact bands. Whole-body adaptation and
    optimization complete the robot's locomotion and manipulation from the
    glasses recording, bridge the embodiment gap between human and humanoid,
    and verify the actions against the reconstruction in physics simulation.
    Visual observations come from re-rendering the reconstructed scene through
    the robot's camera viewpoint along the same rollout, providing paired
    trajectories to train loco-manipulation VLA policies with zero target-task
    real-robot data.}
    \label{fig:overview}
  \end{center}%
  \global\@teasersettrue}
\makeatother

\maketitle

\begin{abstract}
Imitation learning for visuomotor policies typically relies on teleoperated
robot demonstrations, which are costly because every episode requires a trained
operator controlling a physical robot in real time through teleoperation
interfaces and wearable body-tracking hardware.
Egocentric glasses capture human demonstrations without a robot or handheld grippers, yet they track only the
head and in-view human hands, and directly copying human motion to a robot with
different stature, arm reach and hand morphology fails to produce feasible
contacts. \method{} converts egocentric glasses recordings into physically
verified robot training trajectories. An agentic real-to-sim stage reconstructs
collision and textured visual assets from the recorded RGB-D and head streams,
completes unseen surfaces when needed, and predicts target contact bands on the
manipulated objects. During motion adaptation, a grasp planner turns
each contact band into a feasible robot wrist pose and finger closure, bilevel
stance optimization plans collision-free standing postures that keep hand
targets reachable, and a whole-body motion generator synthesizes lower-body
locomotion and out-of-view arm motion. Two-stage motion correction then blends
visibility transitions kinematically and trains a reinforcement-learning
tracking policy in physics simulation under agent-designed semantic rewards,
completing manipulation phases even when the object leaves the camera view.
Rendering the reconstructed scene through the robot's onboard camera along the
optimized rollout yields synchronized robot-view images, proprioception and
verified action labels. Fine-tuned exclusively on this converted dataset,
without real-robot demonstrations, GR00T N1.7 performed loco-manipulation tasks
 autonomously on a Unitree G1. Project page:
\url{https://robonana.github.io/ego2sim2humanoid/}
\end{abstract}

\IEEEpeerreviewmaketitle

\section{Introduction}

\begin{figure*}[t]
  \centering
  \includegraphics[width=0.85\textwidth]{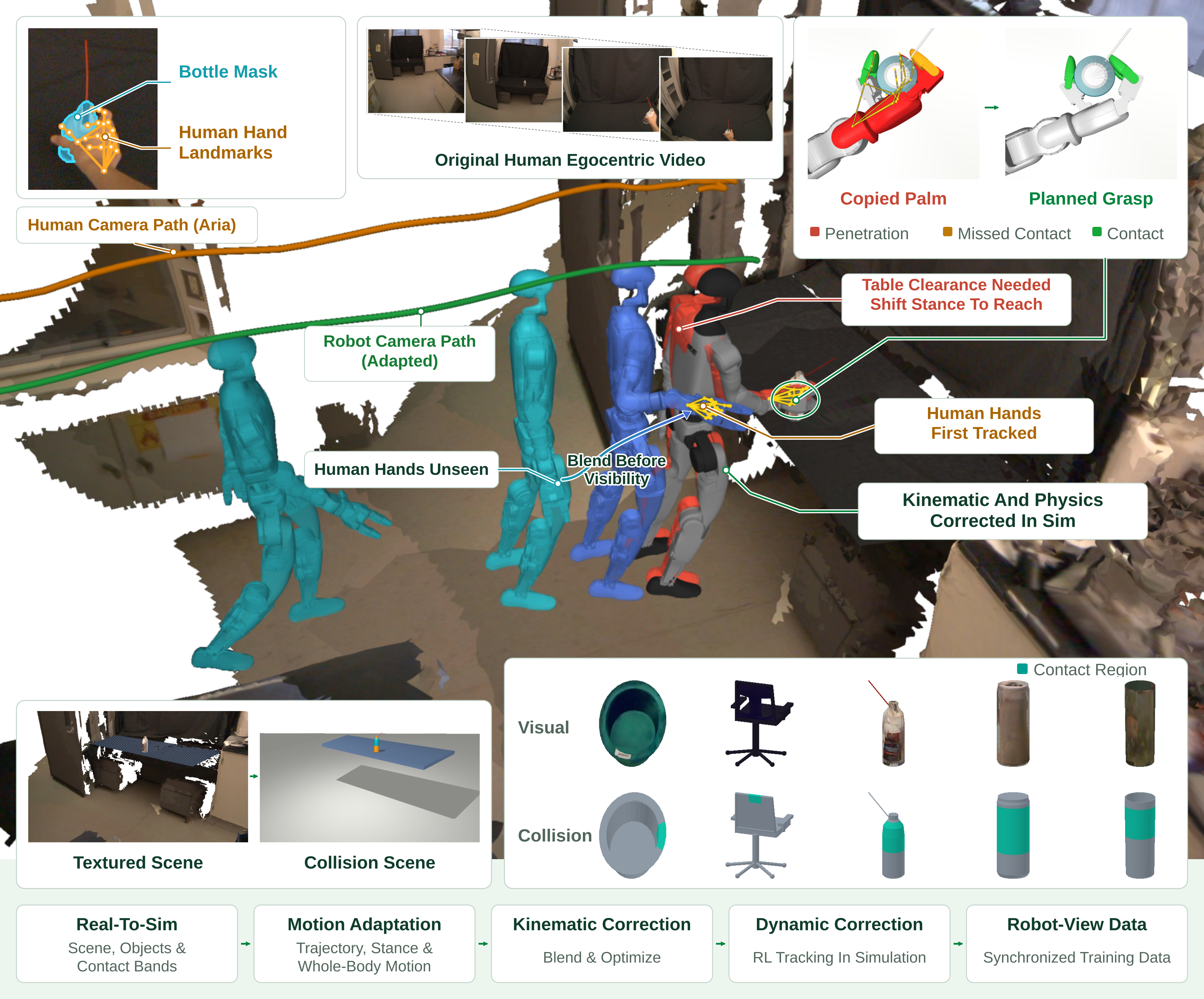}
  \caption{Architecture of \method{}, illustrated on a bottle-lift recording.
  From egocentric video and tracked hand landmarks, agentic real-to-sim masks
  objects and builds a textured visual scene with simplified collision geometry.
  Each object pairs a visual mesh with a collision model marked by target
  contact regions. Along the human (orange) and adapted robot (green) camera
  paths, the robot poses trace adaptation and correction. During the approach
  with hands out of view (cyan), the whole-body motion generator drives
  locomotion and unseen arm motion, then begins blending arm targets prior to
  visibility. Once hands enter the camera view (blue), tracked human landmarks
  take over arm guidance. Placing the robot directly on the projected human
  head path (red) does not guarantee a collision-free stance. Stance adaptation
  is therefore required to keep clear of the table while holding targets within
  reach in a comfortable posture. Directly copying the human palm onto the robot
  hand likewise fails, causing finger penetration or missed contact in the grasp
  comparison above. The final pose (grey) combines bilevel stance adjustment
  with contact-band Dex3 grasp planning. Kinematic optimization and
  reinforcement learning in physics simulation refine this plan into a
  dynamically feasible rollout. \method{} then renders the robot camera view
  from the reconstructed scene along the rollout path to produce synchronized
  training data.}
  \label{fig:contact_motion_adaptation}
\end{figure*}

Whole-body vision-language-action (VLA) models offer a route to generalist
humanoid policies, yet their performance remains bound by the supply of
action-labelled robot trajectories~\cite{groot,pi05,humanoidvla,leverb}.
Internet text and video pretrain visual and semantic representations, but
mapping observations to coordinated joint commands requires physical
demonstrations. Current pipelines collect these demonstrations through
teleoperation across robots and tasks during pretraining and repeat collection
in the target scene for each deployment~\cite{groot,pi05}. Because every
episode requires a physical robot and a trained operator, data collection scales
poorly across new scenes and tasks.

Human recordings provide an alternative data source without a physical robot,
but existing capture setups omit the whole-body motion required for humanoid
loco-manipulation. Egocentric video corpora supply visual and hand-trajectory
pretraining for models such as GR00T N1, EgoVLA, Being-H0 and
H-RDT~\cite{groot,egovla,beingh0,hrdt}. Because an egocentric camera observes
the hands rather than the torso and legs that carry them, policies pretrained on
these videos predict hand and wrist poses and still depend on teleoperated
robot trajectories to ground those poses in whole-body joint
control~\cite{egovla,hrdt}. Portable manipulation interfaces such as UMI and
smart-glasses capture rigs bypass robot teleoperation for fixed-base arms by
placing the robot's gripper or camera on the demonstrator~\cite{umi,datascaling,egozero}.
A handheld gripper or head-mounted camera alone records no stance, footstep,
posture change or balance recovery, separating hand manipulation from the body
motion that positions the hand in space.

Because a humanoid differs from a human demonstrator in height, limb length and
hand morphology, direct retargeting leaves the robot's hand at the wrong
distance from the object, produces infeasible finger contacts, and causes
identical step counts to cover different ground distances. Existing humanoid
transfer pipelines sidestep full-body adaptation rather than resolving this
physical mismatch. Humanoid Policy~$\sim$~Human Policy co-trains a shared
human--humanoid action space but restricts the robot to tabletop manipulation
from a fixed stance~\cite{hat}. EgoHumanoid extends egocentric transfer to
loco-manipulation, yet it reduces lower-body motion to three discrete walking
commands, co-trains with teleoperated robot episodes and translates upper-body
motion kinematically~\cite{egohumanoid}. Such kinematic translation also passes
sensor drift directly into the action labels without verifying whether the
robot maintains balance or achieves stable contact with the object.

\method{} is, to the authors' knowledge, the first pipeline to bring real-to-sim
reconstruction to egocentric data generation for humanoid robots, converting
egocentric recordings into executable whole-body robot
trajectories. The glasses record the head trajectory throughout, the
hands while they are in view, and RGB-D streams of the scene in one metric
frame. For daily-life recordings in the wild without
pre-scanning, an agentic real-to-sim stage reconstructs per-episode assets,
optionally completes unseen surfaces, and predicts target contact bands. Motion
adaptation starts from
the robot's standing pose and projects the head path onto the floor while
planning collision-free stances and upper-body postures so the robot's arms can
reach the interaction targets. Each hand's timeline is split by visibility into
tracked in-view segments, where a rule-based grasp planner maps the predicted
contact band to a feasible robot wrist pose and finger closure, and out-of-view
segments synthesized by a whole-body motion generator. Two-stage motion
correction then blends visibility transitions kinematically and trains a
reinforcement-learning tracking policy for dynamic feasibility in physics simulation under semantic
rewards designed by the agent from the contact events, completing manipulation
phases even when the object moves outside the camera view. The resulting
training label is the closed-loop whole-body action token and hand joint
command that physically moved the simulated object while maintaining balance.

To pair these verified actions with the visual observations the robot would
receive, \method{} renders the reconstructed scene through the robot's onboard
camera along the accepted physics rollout. Each rendered frame shows the
robot's own arms, the object at its simulated pose and the scene from the
robot's camera height. Because images, proprioception and actions come from the
same rollout, vision and control are synchronized without clock alignment. To
demonstrate the fidelity of the converted data directly, whole-body VLA
policies are fine-tuned exclusively on these converted trajectories without any
real-robot demonstrations of the target tasks.

This paper makes five contributions. First, an agentic real-to-sim procedure
reconstructs per-episode collision and textured assets without prior scans,
completes unseen surfaces, and localizes target contact bands on manipulated
objects (\cref{sec:r2s}). Second, a motion
adaptation method combines rule-based grasp planning on the predicted contact
bands with bilevel collision-avoiding stance optimization and whole-body motion
generation across visible and out-of-view hand segments (\cref{sec:gen}).
Third, a two-stage kinematic and reinforcement-learning motion correction
smoothly blends hand visibility transitions and executes the trajectory in
physics simulation under agent-designed semantic rewards that also guide
out-of-view task phases (\cref{sec:corr}). Fourth, robot-view rendering along
the accepted rollout synthesizes synchronized onboard images and verified
action labels (\cref{sec:view}). Fifth, real-robot experiments on a Unitree G1
evaluate policies fine-tuned solely on the converted data without real-robot
demonstrations (\cref{sec:policy,sec:experiments}).

\section{Related Work}
\label{sec:related}

\paragraph{Cross-embodiment and egocentric transfer}
Visual and kinematic transfer methods adapt human or cross-robot recordings by
replacing the visible manipulator and mapping end-effector trajectories, but
they either restrict the robot to a fixed base or leave physical object contact
unverified. Mirage cross-paints a rendered source arm over a target
robot~\cite{mirage}, RoVi-Aug augments arm morphology and camera viewpoint via
diffusion inpainting and view synthesis~\cite{roviaug}, and Phantom and
Masquerade replace human limbs in video with a rendered arm at a kinematically
retargeted pose~\cite{phantom,masquerade}. All four target stationary tabletop
arms without simulating contact, so the overlaid gripper can appear visually
aligned while the command misses the object. EgoMimic co-trains bimanual
tabletop policies on hand-masked egocentric video and teleoperated robot
trajectories~\cite{egomimic}, and Ego2Robot scales kinematic retargeting and
arm rendering across in-the-wild videos for VLA pretraining~\cite{ego2robot}.
EgoEngine~\cite{egoengine} and related video-to-simulation
pipelines~\cite{video2sim2real,egoinfinity,openaoe,human2sim2robot,robopaint} optimize
physically constrained arm trajectories in a digital twin, yet all of them
assume a fixed base rather than a mobile humanoid whose stance adjustments shift
both camera viewpoint and contact timing. EgoHumanoid extends egocentric
transfer to humanoid loco-manipulation, but it decouples the body into three
discrete walking commands and pelvis-relative delta wrist poses, co-trains with
teleoperated robot episodes and translates actions kinematically without
checking grasp stability in physics~\cite{egohumanoid}.

\paragraph{Executable humanoid motion and interaction retargeting}
Tracking retargeted human motion in simulation is standard for whole-body
humanoid control and teleoperation~\cite{omnih2o,humanplus,exbody2,twist,homie},
but policy learning in those systems requires live robot teleoperation, and
free-space retargeting~\cite{gmr} degrades around contact when limb proportions
differ. OKAMI adds object-aware kinematic retargeting from RGB-D video while
keeping the lower body stationary and contact unverified~\cite{okami}.
Prior dexterous grasp synthesis methods optimize hand poses on clean canonical
CAD models without body reachability constraints or scene
context~\cite{grasptta,dexgraspnet}.
OmniRetarget preserves body--terrain--object relationships on an interaction
mesh~\cite{omniretarget}, and HDMI trains reinforcement-learning policies to
co-track retargeted robot and object trajectories from third-person
video~\cite{hdmi}. Both assume a full-body reference aligned with
the scene, whereas egocentric glasses track only the head and, while in view,
the hands, and HDMI only addresses non-prehensile interactions rather than
dexterous grasping. MeshMimic reconstructs terrain from
third-person video to train state-based locomotion policies~\cite{meshmimic},
whereas ULTRA pairs neural retargeting with a multimodal controller trained on
synthetic simulation renders~\cite{ultra}. SONIC provides a quantized whole-body latent action
interface~\cite{sonic}, Kimodo generates motion from sparse
constraints~\cite{kimodo}, and GRAIL combines both with procedural scenes and
video-model trajectories into a synthetic data engine that lacks real scenes
and real human demonstrations~\cite{grail}. 

\paragraph{Concurrent work}
Concurrent human-to-humanoid pipelines~\cite{egoalign,humi,humanashumanoid}
differ in how they verify contact and construct visual inputs. Built on the HumanVerse wearable dataset,
$\lambda_0$ trains a whole-body VLA on robot-aligned human
demonstrations~\cite{lambda0}. It does not
explicitly model the object or scene geometry, which \method{} addresses through
real-scene reconstruction and physics-based interaction validation. EgoHumanoid-V2 combines kinematic reference correction with dynamics-aware controller refinement and robot arm rendering without task-specific robot data~\cite{egohumanoidv2}, but it directly tracks the recorded human palm without adapting for morphology or contact, leaving contact-rich grasping infeasible. 
KODA turns camera and hand references into a whole-body trajectory and synthesizes
onboard views by reprojecting each recorded frame into the robot
camera~\cite{koda}. However, it does not physically validate robot-object interactions in simulation, 
so the generated contacts and motions may not be physically feasible, and it still co-trains
with robot demonstrations of the target tasks.
\method{} executes the adapted motion against a reconstructed, free-moving object in physics simulation and renders the scene directly from the resulting robot rollout, yielding physically
 verified image--action pairs that eliminate the need for real-robot demonstrations.

\section{Method}
\label{sec:method}

\method{} converts human demonstrations recorded with Project Aria Gen~2
glasses~\cite{ariagen2} into physically verified robot trajectories for
whole-body VLA training (\cref{fig:overview}), illustrated on one bottle-lift
recording in \cref{fig:contact_motion_adaptation}. Each recording provides
calibrated RGB-D streams, a gravity-aligned 6-DoF head trajectory, and 3D hand
poses with finger landmarks whenever the hands are in view, without tracking
the torso or legs. Given a recording and task instruction $\ell$, the pipeline
outputs robot training tuples $(I_i, s_i, \ell, a_i)$ at steps $i$, pairing a
rendered robot-camera image $I_i$ and proprioceptive state $s_i$ (43 joint
angles and base-frame gravity) with an action $a_i=(z_i, q^{\text{dex}}_i)$
composed of a 64-D SONIC whole-body latent token $z_i$~\cite{sonic} and 14-D
bimanual Dex3 hand joint commands $q^{\text{dex}}_i$.

\begin{figure*}[t]
  \centering
  \includegraphics[width=\textwidth]{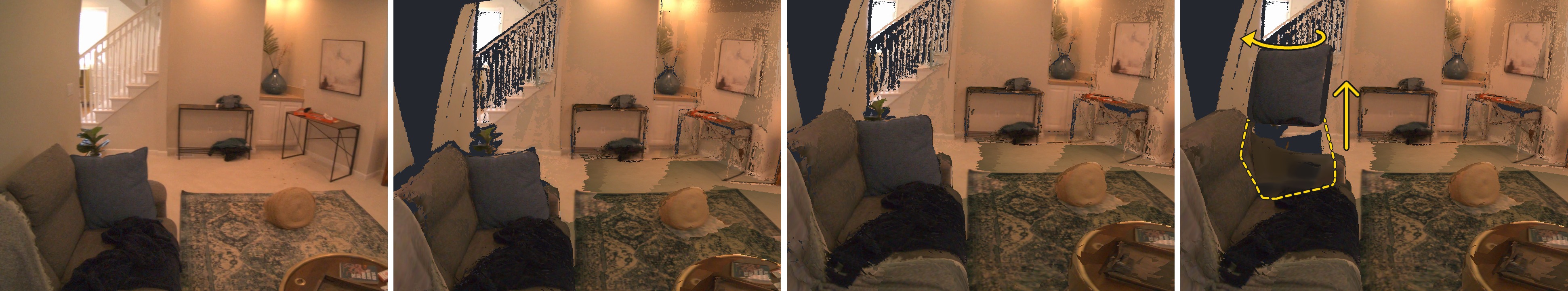}
  \caption{In-the-wild real-to-sim on a home recording from the Aria Gen~2 Pilot Dataset~\cite{ariagen2}. From left: a
  glasses frame; the reconstruction rendered at the recorded camera pose; the
  camera lowered from 1.56~m to a robot camera height of 1.30~m, where surfaces
  the glasses never observed remain dark grey; the pillow, separated as its own
  object with its unseen back completed, lifted 45~cm (straight arrow) and
  turned 20\si{\degree} (curved arrow) from its place on the sofa (dashed
  outline), with the completed sofa
  backing behind it.}
  \label{fig:render}
\end{figure*}

\subsection{Agentic real-to-sim}
\label{sec:r2s}

Device calibration maps the tracked head pose at each frame to the RGB-D
camera, placing the images, stereo depth maps and tracked hands in one
gravity-aligned metric world frame. Rather than pre-scanning a laboratory for
repeated runs, \method{} reconstructs the scene per episode from each daily-life
recording in the wild, with a coding agent running a staged reconstruction procedure 
and checking each stage against source
frames (\cref{app:r2s}) instead of a fixed script. To build the collision scene
for physics simulation,
the agent fits the floor ($z=0$) and support surfaces (such as tables or sofas)
to planes and boundary edges in the stereo depth maps. It then constructs
simplified collision models $\mathcal{M}^{\text{C}}_o$ (boxes, convex parts or
hinged links) for each scene element or object $o$ that preserve the observed
support heights and grasp dimensions. For each moving object $o$, the agent
tracks its 6-DoF pose $T^{\text{obj}}_{o,k}\in\mathrm{SE}(3)$ across frames $k$
by matching its projected silhouette $S_k(u;T^{\text{obj}}_{o,k})\in[0,1]$ to
the object mask $M_{o,k}$ while penalizing overlap with background pixels
$\bar M_k$,
\begin{equation}
  \begin{aligned}
  \min_{\{T^{\text{obj}}_{o,k}\}}\ &\sum_k\Big[\sum_{u\in M_{o,k}}\big(1-S_k(u;T^{\text{obj}}_{o,k})\big)
  +\sum_{u\in\bar M_k}S_k(u;T^{\text{obj}}_{o,k})\Big]\\
  &+\eta\sum_k\big\|T^{\text{obj}}_{o,k+1}\ominus T^{\text{obj}}_{o,k}\big\|^{2},
  \end{aligned}
  \label{eq:track}
\end{equation}
where $A\ominus B$ is the $\mathrm{SE}(3)$ pose difference and $\eta$ weights
smoothness. Hand pixels are excluded from $M_{o,k}$ and $\bar M_k$ so hand
occlusions do not bias the object pose, and frames where an object is fully
occluded follow the contacting body part.

To build the visual meshes for rendering, the agent fuses valid depth maps into
a truncated signed distance field (TSDF)~\cite{curless1996} in each object's
local frame while masking out the demonstrator and moving objects. Multi-view
object masks split the fused surface into separate object and support meshes so
moving an object does not leave a copy on the support, and unoccluded video
frames are projected onto the mesh to bake textures. Each visual mesh is
attached to the same local frame as its collision model $\mathcal{M}^{\text{C}}_o$
so that appearance and contact geometry move together. Because a head-mounted
camera sees only front-facing surfaces and moving an object uncovers the
support beneath it, the agent can optionally complete unseen geometry
(\cref{fig:render}, \cref{app:r2s}). Partly observed compact objects are closed by fitting simple
parametric shapes (such as superellipsoids) to the observed depth and
silhouettes, while supports receive a separate inset backing shell coloured
from nearby visible surfaces. Finally, from the tracked hand and object
trajectories, the agent detects the interaction events and projects candidate object
surfaces into calibrated views around grasp closure to localize a target
contact band in the object's local frame
(\cref{app:r2s}).

\subsection{Motion adaptation}
\label{sec:gen}

Because the glasses record no body posture and the robot differs from the
demonstrator in height, arm reach and hand morphology, human motion cannot be
copied directly. Motion adaptation synthesizes a full-body robot trajectory
$x_{1:N}$ (pelvis pose and joint angles at steps $i=1,\dots,N$) and a monotone
time map $h_i$ from robot step $i$ to human recording time. Adaptation and
correction seek a trajectory starting from the robot's standing posture
$x^{\text{init}}$ that follows the human head path and brings each hand
$a\in\{\text{L},\text{R}\}$ to a target pose $T^{*}_a(h_i)$ whenever visible
($i\in\mathcal{V}_a$), within the set $\mathcal{X}$ of collision-free, balanced
and dynamically executable states,
\begin{equation}
  \begin{aligned}
  \min_{x_{1:N}}\;&\sum_{i}\big\|e^{\text{head}}(x_i,h_i)\big\|_{W_v}^{2}
  +\sum_{a}\sum_{i\in\mathcal{V}_a}\big\|T_{a}(x_i)\ominus T^{*}_{a}(h_i)\big\|_{W_c}^{2}\\
  &\text{s.t.}\ \ x_1=x^{\text{init}},\quad x_i\in\mathcal{X},
  \end{aligned}
  \label{eq:problem}
\end{equation}
where $T_a(x_i)$ is the robot palm pose, $e^{\text{head}}(x_i,h_i)$ measures
planar position, heading and height-scaled vertical error between the robot
camera and human head, and $W_v,W_c$ are weight matrices (\cref{app:adapt}).
Because the human neck moves independently of the torso, the head pose cannot
be rigidly shifted to the pelvis. Projecting the head position onto the floor
plane and smoothing the horizontal viewing direction defines the planar root
reference, with $x^{\text{init}}$ placed at the start (\cref{app:adapt}).

Each hand's timeline is split by camera visibility into tracked in-view
segments and generated out-of-view segments. Within a visible segment, the hand
target $T^{*}_a(h)$ is defined as
\begin{equation}
  T^{*}_a(h)=
  \begin{cases}
  T^{\text{obj}}_o(h)\,T^{\text{grasp}}_a, & \text{during contact with object } o,\\
  T^{\text{hand}}_a(h)\,T^{\text{palm}}, & \text{otherwise},
  \end{cases}
  \label{eq:anchor}
\end{equation}
where $T^{\text{hand}}_a(h)$ is the observed human palm pose and
$T^{\text{palm}}$ maps it to the robot palm frame. During object contact,
directly copying the human palm pose fails because the three-fingered Dex3 hand
differs in morphology and finger geometry from the human hand. Instead, a
rule-based grasp planner turns the predicted contact band on
$\mathcal{M}^{\text{C}}_o$ into a planned wrist offset $T^{\text{grasp}}_a$ and
closed finger posture via bounded least-squares optimization over signed
distances (\cref{app:adapt}), aligning the hand with the contact band, placing
opposing finger pads on opposite sides of the band (or sliding the open hand to
clear the support), preventing link penetration, and keeping palm and forearm
orientations near the human demonstration.

Because the robot's stature and arm reach differ from the demonstrator's,
placing the pelvis on the projected head path can leave $T^{*}_a(h_n)$
unreachable or cause scene collisions. \method{} therefore plans collision-free
planar stances $s=(p_s,\psi_s)\in\mathrm{SE}(2)$ together with waist and arm
postures $y_n=(\theta^{\text{w}}_n,\theta^{\text{arm}}_n)$ at interaction
keyframes $h_n$ via bilevel optimization. Given a candidate stance $s$, the
inner problem solves for $y_n(s)$ within joint limits $\Theta$,
\begin{equation}
  \begin{aligned}
  y_n(s)=\arg\min_{y\in\Theta}\ &\sum_{a}\big\|T_{a}(s,y)\ominus T^{*}_{a}(h_n)\big\|_{W_a}^{2}\\
  &+\big\|e^{\text{head}}(s,y;h_n)\big\|_{W_v}^{2}
  +\big\|y-y^{\text{stand}}\big\|_{W_y}^{2},
  \end{aligned}
  \label{eq:keyframe}
\end{equation}
with $W_a\gg W_v$ so reaching the hand target takes priority and the remaining
waist degrees of freedom lean toward the human head pose. Writing $J_n(s)$ for
the minimum of \cref{eq:keyframe}, the outer problem selects a stance $s^{*}$
that clears scene obstacles $\mathcal{O}$ by margin $d_{\text{s}}$,
\begin{equation}
  s^{*}=\arg\min_{s\in\mathrm{SE}(2)}\ \sum_n J_n(s)\quad
  \text{s.t.}\ \ d(s;\mathcal{O})\ge d_{\text{s}}.
  \label{eq:stance}
\end{equation}
After stretching waypoint durations to stay within robot speed limits, a
whole-body motion generator~\cite{kimodo} conditioned on the root path, visible
hand targets and planned stances $(s^{*},y_n(s^{*}))$ synthesizes $\hat x_{1:N}$
across overlapping windows (\cref{app:adapt}).

\begin{figure*}[t]
  \centering
  \includegraphics[height=125pt]{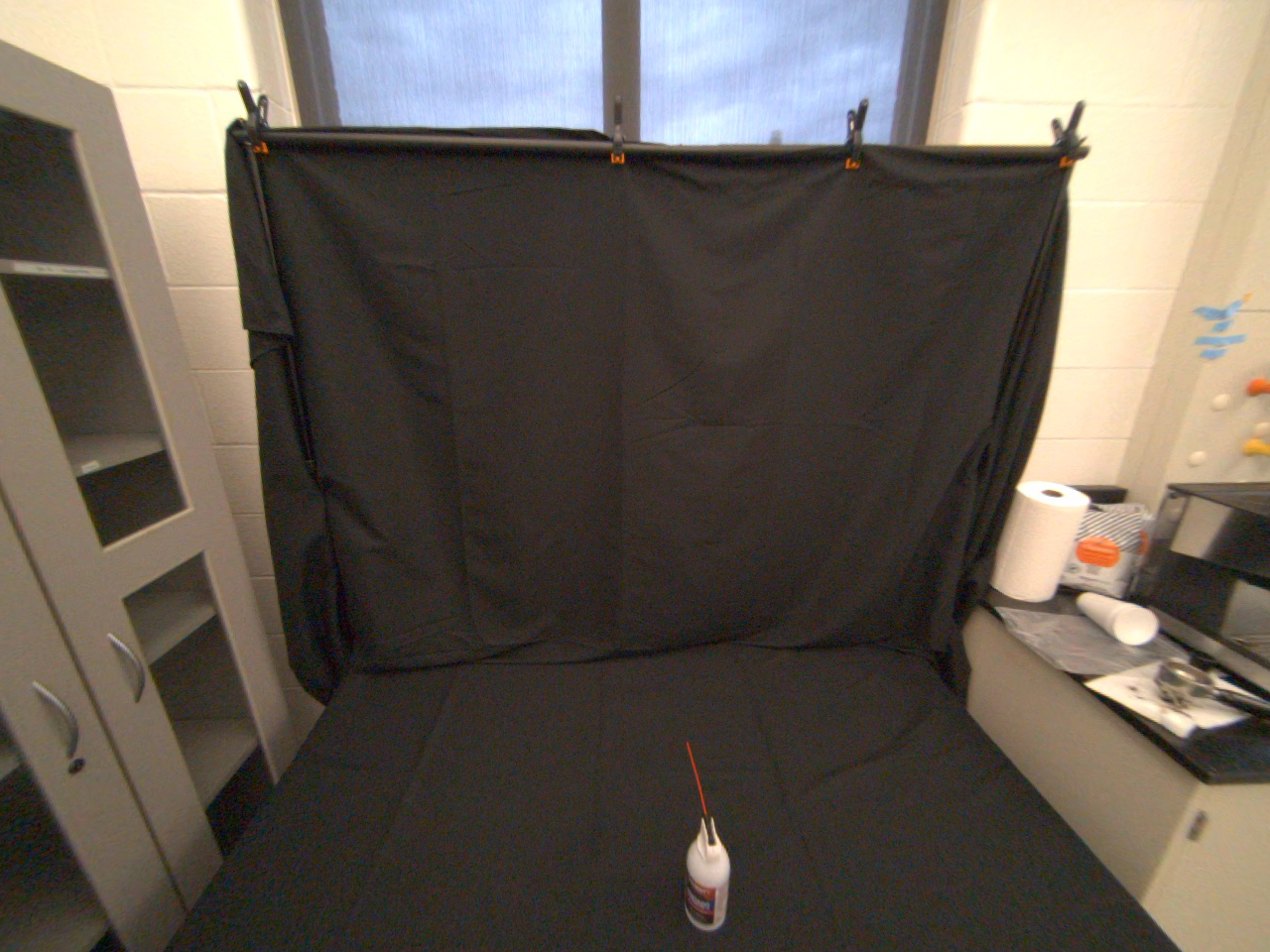}\hfill
  \includegraphics[height=125pt]{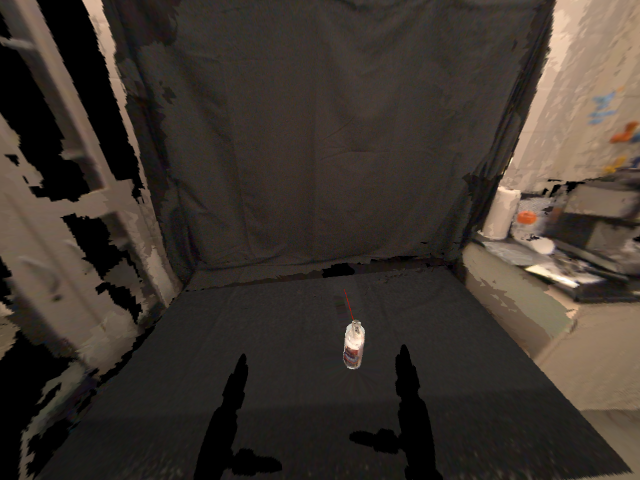}\hfill
  \includegraphics[height=125pt,trim=0 0 0 6,clip]{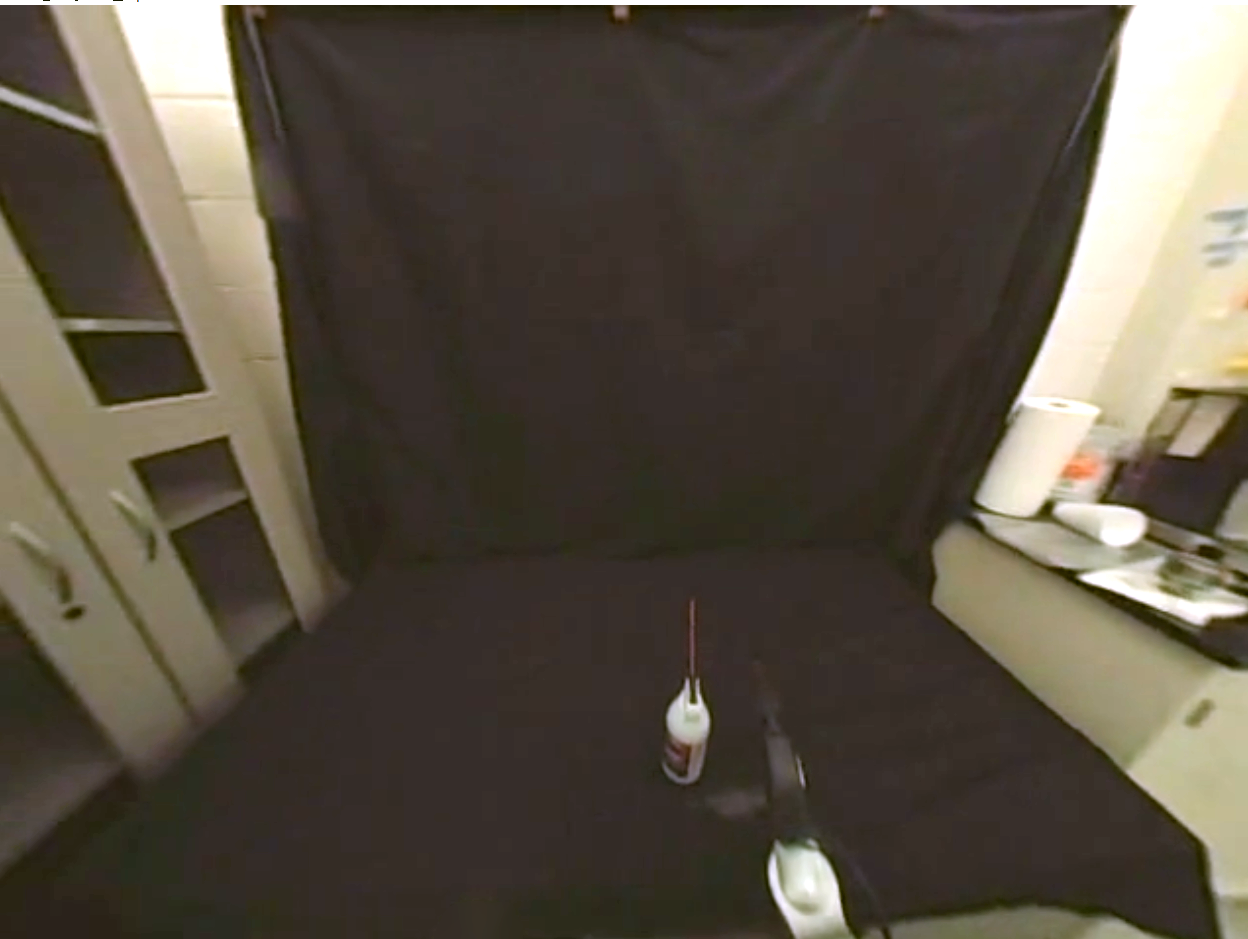}
  \caption{Views of the bottle-lift scene. From left: a frame from the glasses;
  the simulated robot view, rendered from the reconstruction at the robot
  camera as used for training; the real robot view from the G1 head camera
  during a real-robot trial. The rendered view takes the robot's viewpoint and
  arms instead of the person's, while colour balance and regions the glasses
  never observed still differ from the real camera.}
  \label{fig:views}
\end{figure*}

\subsection{Kinematic motion correction}
\label{sec:corr}

Because the motion generator meets keyframes only approximately and leaves
out-of-view hands unconstrained, switching directly to the tracked hand when it
enters view would cause a velocity jump, and $\hat x_{1:N}$ can collide with
the scene or push the object before grasp closure. The hand target therefore
blends the generated hand pose into the tracked target over short fade windows
around each visibility transition (\cref{app:corr}), and the grasp pose is
re-planned on the generated body before trajectory optimization refines
$\hat x_{1:N}$ to track the hand targets, prevent foot slip, keep the centre of
mass in the support polygon, and clear scene collisions. The wrist target stays
fixed in the object frame from grasp closure to lift-off while closing to
$q^{\text{dex}}_i$, and a kinematic physics replay checks that the hand does
not push the object before closure.

\subsection{Dynamic correction via reinforcement learning}

Kinematic planning fixes the coarse hand placement and approach direction, but
cannot ensure balance or grasp retention under gravity, or guide phases where
the object leaves the camera view. \method{} therefore trains a tracking policy
$\pi_\phi$ by reinforcement learning in a physics simulation of the
reconstructed scene with free-moving objects. At step $i$, $\pi_\phi$ maps
$s_i$ and upcoming reference states $\hat x_{i:i+K}$ to a 64-D token $z_i$, the
frozen SONIC decoder $D$~\cite{sonic} outputs joint targets
$\theta^{\text{tgt}}_i$, and the simulator $\mathcal{F}$ steps $s_i$ and the
object poses $\chi_i$,
\begin{equation}
  \begin{aligned}
  z_i&=\pi_\phi\big(s_i,\hat x_{i:i+K}\big),\quad
  \theta^{\text{tgt}}_i=D\big(z_i,s_i\big),\\
  (s_{i+1},\chi_{i+1})&=\mathcal{F}\big(s_i,\chi_i,\theta^{\text{tgt}}_i,q^{\text{dex}}_i\big).
  \end{aligned}
  \label{eq:rollout}
\end{equation}
Rewards and early terminations are instantiated automatically from per-task semantics, detected contact events and
predicted contact regions (\cref{app:reward}), combining body-tracking and
token-smoothness terms with contact, object-tracking and support-clearance
terms during manipulation. When a task phase moves outside
the camera view (such as placing a hat on the head), semantic rewards specify
the target object-to-body relation so reinforcement learning completes the task
without camera observations, while terminating on falls, collisions, lost
contact or object drift. Rollouts are accepted only if the object reaches its
goal and head tracking error and joint motions stay within physical bounds
(\cref{app:corr}), yielding the verified closed-loop action label
$a_i=(z_i, q^{\text{dex}}_i)$.

\subsection{Robot-view rendering and policy learning}
\label{sec:view}
\label{sec:policy}

To avoid the human viewpoint, hands and timing of glasses video, \method{}
renders each training image $I_i$ through the robot camera along the accepted
rollout (\cref{fig:render}, \cref{app:render}). The renderer composites the
static background, dynamic objects at simulated poses $\chi_i$, and robot links
at state $s_i$. Segmenting moving objects during reconstruction (\cref{sec:r2s})
avoids duplicate objects and places simulated objects at exact physics poses.
Unobserved areas retain median scene colour, while robot links take physical arm
and Dex3 materials. \Cref{fig:views} places a glasses frame beside a rendered
training view and the real robot camera in the bottle-lift scene. Because $I_i$, $s_i$ and $a_i$ come from the same rollout,
vision and control are synchronized by construction. To demonstrate the fidelity of the converted
data, GR00T N1.7 (3B)~\cite{groot17} is fine-tuned only on these tuples
$(I_i, s_i, \ell, a_i)$ without real-robot data to predict 78-D action chunks
$a_i$, decoded on the G1 by the frozen SONIC decoder~\cite{sonic}.

\section{Experiments}
\label{sec:experiments}

\begin{figure*}[t]
  \centering
  \includegraphics[height=103pt,trim=262 0 261 0,clip]{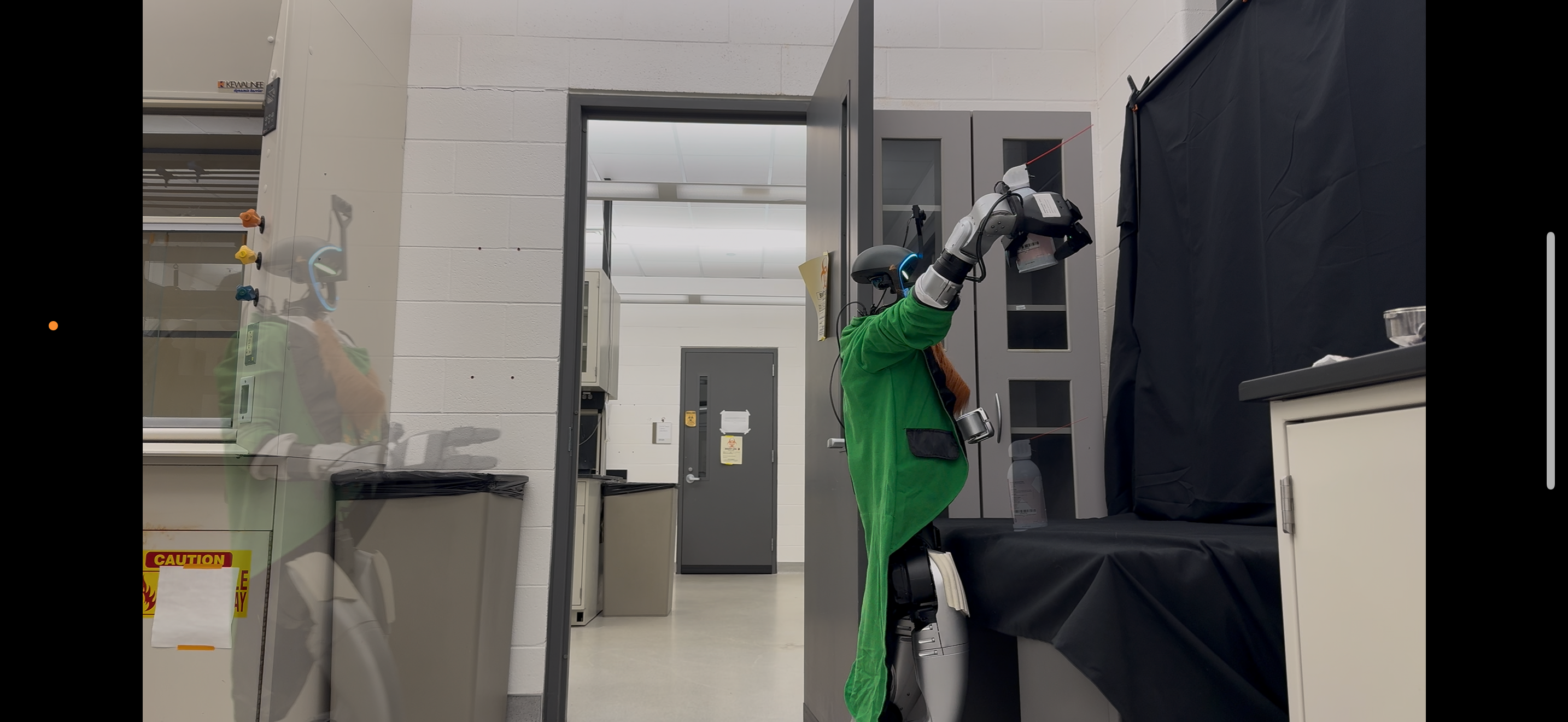}\hspace{8pt}%
  \includegraphics[height=103pt,trim=262 0 261 0,clip]{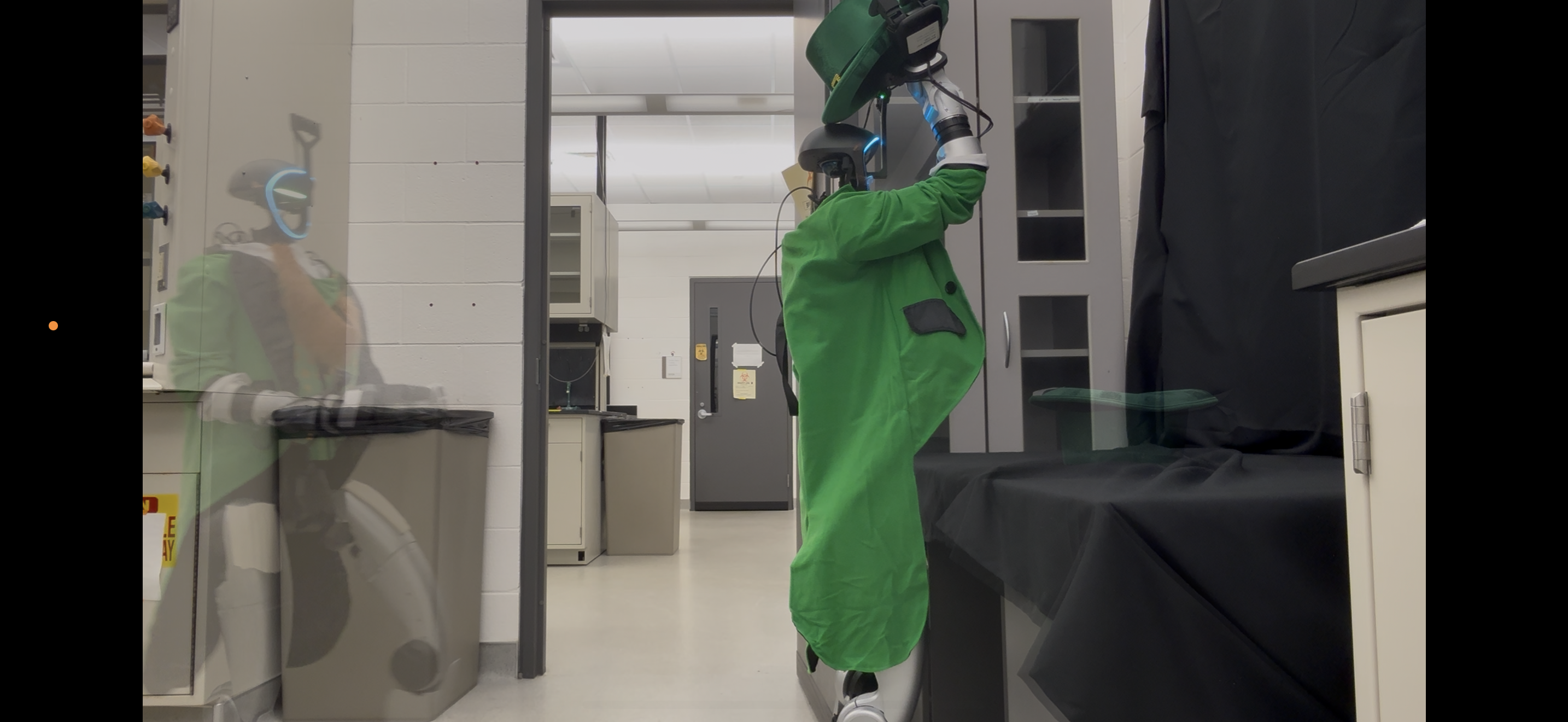}
  \caption{Real-robot trials of bottle lift (left) and hat wearing (right), each
  overlaying earlier frames of the trial as translucent images.}
  \label{fig:real}
\end{figure*}

\paragraph{Tasks and platform}
Experiments evaluate trajectory conversion in simulation and a GR00T~N1.7
policy trained solely on the converted dataset, using no real-robot
demonstrations of the target tasks. Human demonstrations were collected without
robot hardware or body trackers by wearing only Aria Gen~2 glasses while
walking and performing each task, capturing calibrated RGB-D
streams, a gravity-aligned 6-DoF head trajectory, and in-view 3D hand poses.
In total, 219 episodes were collected for \emph{bottle lift} (walk to a table,
grasp a bottle with the right hand, and lift it) and 154 for \emph{hat wearing}
(walk to a table, pick up a hat, and put it on the head), each paired with one
language instruction. Hardware evaluations run on a 29-DoF Unitree G1 humanoid
equipped with Dex3 hands and a head camera.

\paragraph{Data filtering}
A recording enters the training set only if it passes every stage of the
pipeline: motion adaptation, the kinematic physics replay, the physics rollout
check and a final review of the assembled episodes. Of the 219 bottle-lift
recordings, 189 enter the training set, compared with 59 of the 154
hat-wearing recordings, most of which are lost before simulation. The main
reason is a grasp that touches the hat without holding it. The grasp planner
places the finger pads on the hat wall inside the contact band, but the wall is
only millimetres thick, which leaves the plan almost no margin. In some
recordings the thumb inside the crown stays clear of the wall instead of
opposing the fingers, and in others the hand drifts by a few millimetres
between closure and lift-off. The kinematic physics replay closes the hand on
the free-moving hat, finds that contact does not last, and the episode is
discarded rather than edited to fit. The bottle's thick body gives the pads
more margin, so bottle lift rarely fails this way.

\paragraph{Grasp planning and correction}
Planning the grasp against the predicted contact band reduces hand penetration
and stabilizes the hat during manipulation compared with copying the retargeted
human palm pose (\cref{tab:contact}). Across the 71 checked hat recordings,
the copied-palm baseline places the robot wrist at the retargeted human palm
and closes the hand into a fixed posture at the recorded contact time while
leaving all other stages unchanged, and both references pass through the same
reinforcement-learning stage. Penetration is the deepest intrusion of any hand
link into the hat between hand closure and release, and in-hand motion is the
largest translation and rotation of the hat in the hand frame while held, until
the hat reaches the head. The copied palm drove the thumb and index finger
deeper into the hat wall than the planned grasp in 65 of 71 recordings,
raising median penetration from 0.8 to 1.8\,cm. Across the 43 recordings that
succeeded under both conditions, the hat moved further inside the copied-palm
grasp in 36 cases (Wilcoxon $p < 0.001$ for both comparisons), with median
in-hand shift rising from 3.7 to 6.6\,cm and rotation from \SI{22.9}{\degree}
to \SI{33.0}{\degree}.

This difference arises from hand morphology. Placing the three Dex3 fingers at
human finger locations taken from the five-fingered human palm drives the thumb
or index finger through the thin hat wall, whereas the planner positions the
finger pads on opposite sides of the predicted contact band. Contacts still fell
inside the predicted band in both conditions because the demonstrator gripped
the rim that the band marks, and no slip occurred in either kinematic reference
because both keep the hat attached to the closed hand. In-hand motion
therefore appeared only once reinforcement learning tracked the reference
under contact physics. Because 56 of 71 planned rollouts succeeded compared
with 53 for the copied palm (with 43 succeeding in both conditions and 23 in
only one), the grasp planner benefits contact quality rather than raw success
count.

\begin{table}[t]
  \centering
  \caption{Human palm pose copied to the robot wrist (Human) against the grasp
  planner (Planner) on 71 hat recordings, each followed by the same
  reinforcement-learning stage. Median with interquartile range, reporting
  penetration over all recordings and in-hand motion over successful rollouts
  until the hat reaches the head.}
  \label{tab:contact}
  \begin{tabular}{lcc}
    \toprule
    & Human & Planner \\
    \midrule
    Penetration (cm) & 1.8 [1.7, 2.0] & 0.8 [0.4, 1.4] \\
    In-hand shift (cm) & 6.6 [5.3, 7.4] & 3.7 [2.6, 5.4] \\
    In-hand rotation (\si{\degree}) & 33.0 [29.1, 41.0] & 22.9 [18.1, 26.6] \\
    \bottomrule
  \end{tabular}
\end{table}

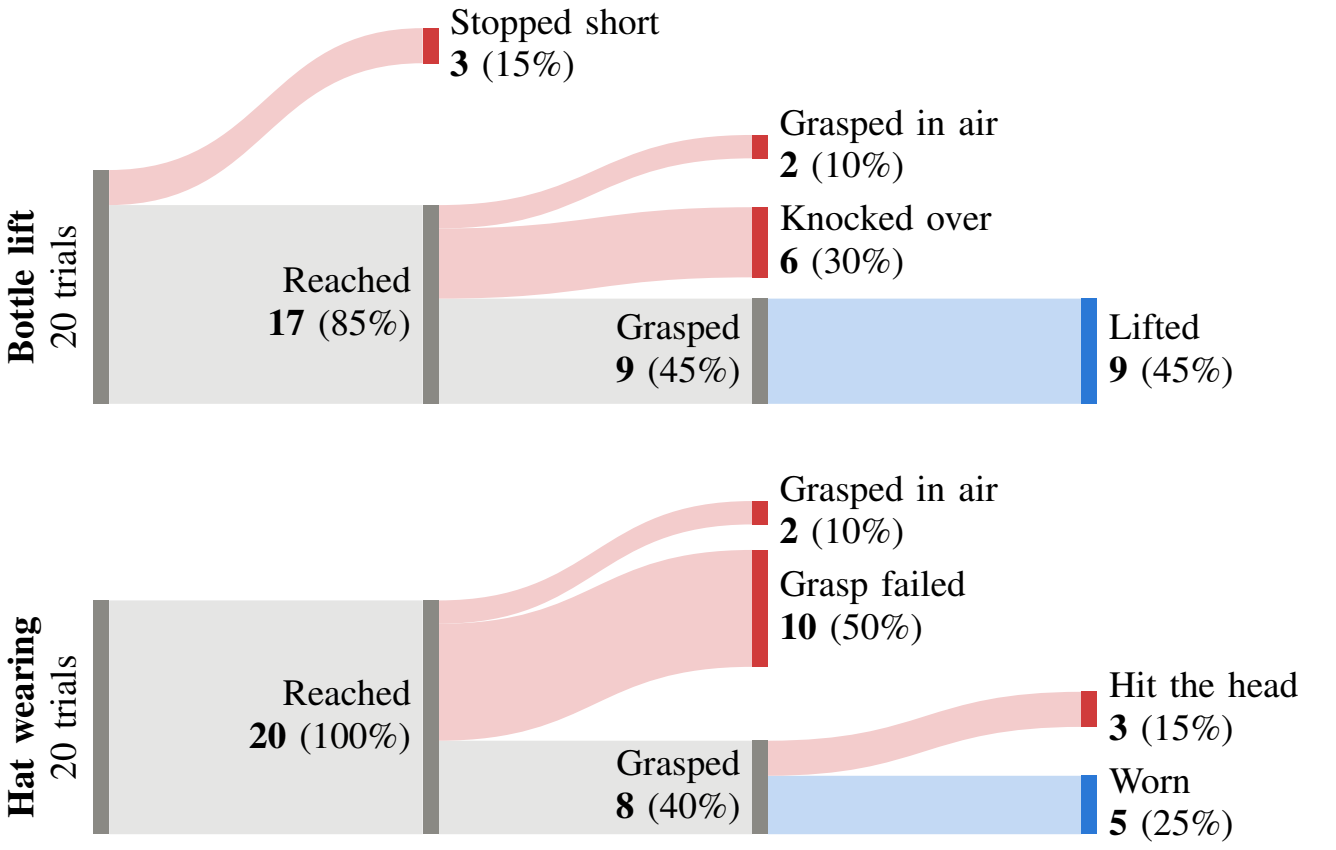
\begin{figure}[t]
  \centering
  \definecolor{ffGrey}{HTML}{8A8984}\definecolor{ffBlue}{HTML}{2A78D6}\definecolor{ffRed}{HTML}{D03B3B}
\colorlet{ffGreyFlow}{ffGrey!22!white}\colorlet{ffBlueFlow}{ffBlue!28!white}\colorlet{ffRedFlow}{ffRed!26!white}
\begin{tikzpicture}[x=1pt,y=1pt,font=\scriptsize,
  lbl/.style={inner sep=0pt,align=left,anchor=west},
  lblL/.style={inner sep=0pt,align=right,anchor=east}]
\fill[ffGreyFlow] (17.0,0.0) .. controls (46.5,0.0) and (46.5,0.0) .. (76.0,0.0) -- (76.0,44.0) .. controls (46.5,44.0) and (46.5,44.0) .. (17.0,44.0) -- cycle;
\fill[ffGreyFlow] (79.0,0.0) .. controls (108.5,0.0) and (108.5,0.0) .. (138.0,0.0) -- (138.0,17.6) .. controls (108.5,17.6) and (108.5,17.6) .. (79.0,17.6) -- cycle;
\fill[ffRedFlow] (79.0,17.6) .. controls (108.5,17.6) and (108.5,31.5) .. (138.0,31.5) -- (138.0,53.5) .. controls (108.5,53.5) and (108.5,39.6) .. (79.0,39.6) -- cycle;
\fill[ffRedFlow] (79.0,39.6) .. controls (108.5,39.6) and (108.5,58.3) .. (138.0,58.3) -- (138.0,62.7) .. controls (108.5,62.7) and (108.5,44.0) .. (79.0,44.0) -- cycle;
\fill[ffBlueFlow] (141.0,0.0) .. controls (170.5,0.0) and (170.5,0.0) .. (200.0,0.0) -- (200.0,11.0) .. controls (170.5,11.0) and (170.5,11.0) .. (141.0,11.0) -- cycle;
\fill[ffRedFlow] (141.0,11.0) .. controls (170.5,11.0) and (170.5,20.2) .. (200.0,20.2) -- (200.0,26.8) .. controls (170.5,26.8) and (170.5,17.6) .. (141.0,17.6) -- cycle;
\fill[ffGrey] (14.0,0.0) rectangle (17.0,44.0);
\fill[ffGrey] (76.0,0.0) rectangle (79.0,44.0);
\node[lblL] at (74.0,22.0) {Reached\\\textbf{20} (100\%)};
\fill[ffGrey] (138.0,0.0) rectangle (141.0,17.6);
\node[lblL] at (136.0,8.8) {Grasped\\\textbf{8} (40\%)};
\fill[ffBlue] (200.0,0.0) rectangle (203.0,11.0);
\node[lbl] at (205.0,5.5) {Worn\\\textbf{5} (25\%)};
\fill[ffRed] (200.0,20.2) rectangle (203.0,26.8);
\node[lbl] at (205.0,23.5) {Hit the head\\\textbf{3} (15\%)};
\fill[ffRed] (138.0,31.5) rectangle (141.0,53.5);
\node[lbl] at (143.0,42.5) {Grasp failed\\\textbf{10} (50\%)};
\fill[ffRed] (138.0,58.3) rectangle (141.0,62.7);
\node[lbl] at (143.0,60.5) {Grasped in air\\\textbf{2} (10\%)};
\node[rotate=90,anchor=south,inner sep=0pt,align=center] at (11.0,22.0) {\textbf{Hat wearing}\\20 trials};
\fill[ffGreyFlow] (17.0,81.0) .. controls (46.5,81.0) and (46.5,81.0) .. (76.0,81.0) -- (76.0,118.4) .. controls (46.5,118.4) and (46.5,118.4) .. (17.0,118.4) -- cycle;
\fill[ffRedFlow] (17.0,118.4) .. controls (46.5,118.4) and (46.5,145.1) .. (76.0,145.1) -- (76.0,151.7) .. controls (46.5,151.7) and (46.5,125.0) .. (17.0,125.0) -- cycle;
\fill[ffGreyFlow] (79.0,81.0) .. controls (108.5,81.0) and (108.5,81.0) .. (138.0,81.0) -- (138.0,100.8) .. controls (108.5,100.8) and (108.5,100.8) .. (79.0,100.8) -- cycle;
\fill[ffRedFlow] (79.0,100.8) .. controls (108.5,100.8) and (108.5,104.8) .. (138.0,104.8) -- (138.0,118.0) .. controls (108.5,118.0) and (108.5,114.0) .. (79.0,114.0) -- cycle;
\fill[ffRedFlow] (79.0,114.0) .. controls (108.5,114.0) and (108.5,127.2) .. (138.0,127.2) -- (138.0,131.6) .. controls (108.5,131.6) and (108.5,118.4) .. (79.0,118.4) -- cycle;
\fill[ffBlueFlow] (141.0,81.0) .. controls (170.5,81.0) and (170.5,81.0) .. (200.0,81.0) -- (200.0,100.8) .. controls (170.5,100.8) and (170.5,100.8) .. (141.0,100.8) -- cycle;
\fill[ffGrey] (14.0,81.0) rectangle (17.0,125.0);
\fill[ffGrey] (76.0,81.0) rectangle (79.0,118.4);
\node[lblL] at (74.0,99.7) {Reached\\\textbf{17} (85\%)};
\fill[ffGrey] (138.0,81.0) rectangle (141.0,100.8);
\node[lblL] at (136.0,90.9) {Grasped\\\textbf{9} (45\%)};
\fill[ffBlue] (200.0,81.0) rectangle (203.0,100.8);
\node[lbl] at (205.0,90.9) {Lifted\\\textbf{9} (45\%)};
\fill[ffRed] (138.0,104.8) rectangle (141.0,118.0);
\node[lbl] at (143.0,111.4) {Knocked over\\\textbf{6} (30\%)};
\fill[ffRed] (138.0,127.2) rectangle (141.0,131.6);
\node[lbl] at (143.0,129.4) {Grasped in air\\\textbf{2} (10\%)};
\fill[ffRed] (76.0,145.1) rectangle (79.0,151.7);
\node[lbl] at (81.0,148.4) {Stopped short\\\textbf{3} (15\%)};
\node[rotate=90,anchor=south,inner sep=0pt,align=center] at (11.0,103.0) {\textbf{Bottle lift}\\20 trials};
\end{tikzpicture}
  \caption{Outcomes of the 20 real-robot trials per task. Bands follow the
  trials from the start through reaching the table and the grasp to the lift or
  wear, band height is proportional to the number of trials, and failures leave
  the band at the stage where they occurred. Grasped in air indicates that the
  hand closed away from the object, and percentages are of all trials of the
  task.}
  \label{fig:failures}
\end{figure}

\paragraph{Comparison with EgoHumanoid}
On the same hat recordings, EgoHumanoid~\cite{egohumanoid} and
EgoHumanoid-V2~\cite{egohumanoidv2} traversed only 0.5 to 0.6 of the recorded
path in median ratio, left the hand a median of 70.8 to 112.9\,cm from the
contact region at touch time, and admitted zero feasible rim grasps, whereas
\method{} achieved a path ratio of 0.9, placed the hand at a median of
0.8\,cm, and yielded a feasible grasp in 57 of 71 recordings (\cref{tab:ego}).
Both baselines were run from their released code, with EgoHumanoid and
kinematic EgoHumanoid-V2 evaluated on all 71 recordings and full
EgoHumanoid-V2 (with its dynamics refinement) evaluated on the 6 of 8 held-out
recordings it completed. Hand-to-contact distance was measured at each
rollout's own pose at the recorded touch time. Feasible rim grasps were then
tested with each output wrist pose fixed, and with the baseline robots moved to
where the demonstrator stood, by solving the fingers of the EgoHumanoid-V2
robot hand to hook the hat rim without penetrating it. Even with walking error
removed in this way, neither baseline yielded a feasible rim grasp in any
recording.

Both baselines replay the demonstrator's body and palm motion without a model
of the scene or the object. Given the difference in morphology between the
humanoid robot and the human demonstrator, walking and hand-pose errors
compound. Even after moving a baseline robot to the demonstrator's standing position, a
palm pose matched to the human hand does not yield successful contact, as the
full EgoHumanoid-V2 palm lay 6\,cm from the human palm while the solved
fingers stopped 2 to 6\,cm short of the contact band. \method{} instead plans full-body motion against the reconstructed scene and hand contact poses on the predicted band.

\begin{table}[t]
  \centering
  \caption{EgoHumanoid baselines against \method{} on the hat recordings.
  $n$ is the number of recordings, path the median ratio of robot to human
  path length, hand the median distance from the hand to the contact region at
  the recorded touch time, and grasps the number of feasible rim grasps found
  after moving the baseline robots to where the demonstrator stood.}
  \label{tab:ego}
  \begin{tabular}{lcccc}
    \toprule
    & $n$ & Path & Hand (cm) & Grasps \\
    \midrule
    EgoHumanoid & 71 & 0.5 & 70.8 & 0 \\
    EgoHumanoid-V2, kinematic & 71 & 0.6 & 93.9 & 0 \\
    EgoHumanoid-V2, full & 6 & 0.6 & 112.9 & 0 \\
    \method{} & 71 & 0.9 & 0.8 & 57 \\
    \bottomrule
  \end{tabular}
\end{table}

\paragraph{Real-robot results}
Fine-tuned solely on the converted dataset with no target-task real-robot data,
the GR00T~N1.7 policy lifted the bottle in 9 of 20 autonomous trials (45\%) and
put the hat on in 5 of 20 (25\%) (\cref{fig:real,fig:failures}). On bottle
lift, 3 trials (15\%) stopped short of the table and 17 (85\%) reached it. Of
those 17 reaching trials, 8 failed during grasping (40\% of all 20 trials, or
47\% of reaching trials), either by knocking the bottle over (6, 30\%) or by
closing the hand in the air (2, 10\%), whereas all 9 grasped bottles (45\%)
were lifted. On hat wearing, all 20 trials reached the table (100\%), and 12
(60\%) failed during grasping, 10 failing to grasp the hat (50\%) and 2 closing
the hand in the air (10\%). Among the 8 trials (40\%) that grasped the hat, 3
(38\% of grasps, or 15\% of all trials) hit the robot's head while placing the
hat outside the head camera's view, leaving 5 successful placements (25\%).

\section{Limitations}

Physical dynamics parameters of reconstructed objects, including mass, inertia
and surface friction, are estimated from priors by the coding agent from video. 
Although reinforcement-learning adaptation tolerates
moderate parameter variation on lightweight household items, manipulating heavy
objects or payloads with uneven mass distributions can still suffer from a
sim-to-real dynamics gap. Representing objects as rigid bodies, parametric
shapes or hinged links also limits the real-to-sim stage on highly deformable
or topology-changing tasks such as folding clothes or cutting bread. 

\section{Conclusion}
\label{sec:conclusion}

A recording from egocentric glasses captures where the person looked and,
while in view, what the hands did. By localizing target contact bands on
reconstructed objects, planning feasible robot grasps and collision-avoiding
stances, executing the trajectory in physics simulation under semantic RL
rewards, and rendering the scene from the robot's onboard camera along that same
rollout, \method{} converts non-embodied human recordings into physically
grounded training data. Fine-tuning a whole-body VLA model solely on these
converted demonstrations without real-robot data achieves autonomous
loco-manipulation on a Unitree G1. With no robot hardware, trained operators or
separate pre-scans required during recording, \method{} moves toward the
ultimate goal of scalable data collection for humanoid loco-manipulation that is
accessible to anyone in any scene.

\clearpage
\bibliographystyle{plainnat-etal}
\bibliography{references}

\begin{thebibliography}{49}
\providecommand{\natexlab}[1]{#1}
\providecommand{\url}[1]{\texttt{#1}}
\expandafter\ifx\csname urlstyle\endcsname\relax
  \providecommand{\doi}[1]{doi: #1}\else
  \providecommand{\doi}{doi: \begingroup \urlstyle{rm}\Url}\fi

\bibitem[Araujo et~al.(2025)Araujo, Ze, Xu, Wu, and Liu]{gmr}
Joao~Pedro Araujo, Yanjie Ze, Pei Xu, Jiajun Wu, and C.~Karen Liu.
\newblock Retargeting matters: General motion retargeting for humanoid motion
  tracking.
\newblock \emph{arXiv preprint arXiv:2510.02252}, 2025.

\bibitem[Ben et~al.(2025)Ben, Jia, Zeng, Dong, Lin, and Pang]{homie}
Qingwei Ben, Feiyu Jia, Jia Zeng, Junting Dong, Dahua Lin, and Jiangmiao Pang.
\newblock {HOMIE}: Humanoid loco-manipulation with isomorphic exoskeleton
  cockpit.
\newblock \emph{arXiv preprint arXiv:2502.13013}, 2025.

\bibitem[Bi et~al.(2025)Bi, Wu, Lin, Tan, et~al.]{hrdt}
Hongzhe Bi, Lingxuan Wu, Tianwei Lin, Hengkai Tan, et~al.
\newblock {H-RDT}: Human manipulation enhanced bimanual robotic manipulation.
\newblock \emph{arXiv preprint arXiv:2507.23523}, 2025.

\bibitem[Chen et~al.(2026)Chen, Jiang, Xu, et~al.]{egohumanoidv2}
Jin Chen, Yiming Jiang, Chongyang Xu, et~al.
\newblock {EgoHumanoid-V2}: Human-to-humanoid transfer of coordinated
  whole-body skills for loco-manipulation.
\newblock \emph{arXiv preprint arXiv:2609.37181}, 2026.

\bibitem[Chen et~al.(2024{\natexlab{a}})Chen, Hari, Dharmarajan, Xu, Vuong, and
  Goldberg]{mirage}
Lawrence~Yunliang Chen, Kush Hari, Karthik Dharmarajan, Chenfeng Xu, Quan
  Vuong, and Ken Goldberg.
\newblock Mirage: Cross-embodiment zero-shot policy transfer with
  cross-painting.
\newblock In \emph{Robotics: Science and Systems (RSS)}, 2024{\natexlab{a}}.
\newblock arXiv:2402.19249.

\bibitem[Chen et~al.(2024{\natexlab{b}})Chen, Xu, Dharmarajan, Irshad, Cheng,
  Keutzer, Tomizuka, Vuong, and Goldberg]{roviaug}
Lawrence~Yunliang Chen, Chenfeng Xu, Karthik Dharmarajan, et~al.
\newblock {RoVi-Aug}: Robot and viewpoint augmentation for cross-embodiment
  robot learning.
\newblock In \emph{Conference on Robot Learning (CoRL)}, 2024{\natexlab{b}}.
\newblock arXiv:2409.03403.

\bibitem[Chi et~al.(2024)Chi, Xu, Pan, Cousineau, Burchfiel, Feng, Tedrake, and
  Song]{umi}
Cheng Chi, Zhenjia Xu, Chuer Pan, et~al.
\newblock Universal manipulation interface: In-the-wild robot teaching without
  in-the-wild robots.
\newblock In \emph{Robotics: Science and Systems (RSS)}, 2024.

\bibitem[Curless and Levoy(1996)]{curless1996}
Brian Curless and Marc Levoy.
\newblock A volumetric method for building complex models from range images.
\newblock In \emph{Proc. SIGGRAPH}, pages 303--312, 1996.

\bibitem[Ding et~al.(2025)Ding, Ma, Tong, Zou, et~al.]{humanoidvla}
Pengxiang Ding, Jianfei Ma, Xinyang Tong, Binghong Zou, et~al.
\newblock {Humanoid-VLA}: Towards universal humanoid control with visual
  integration.
\newblock \emph{arXiv preprint arXiv:2502.14795}, 2025.

\bibitem[Fan et~al.(2026)Fan, Zhao, Zhang, Chen, Wang, Zhang, and
  Cheng]{robopaint}
Jiacheng Fan, Zhiyue Zhao, Yiqian Zhang, et~al.
\newblock {RoboPaint}: From human demonstration to any robot and any view.
\newblock \emph{arXiv preprint arXiv:2602.05325}, 2026.

\bibitem[Fu et~al.(2024)Fu, Zhao, Wu, Wetzstein, and Finn]{humanplus}
Zipeng Fu, Qingqing Zhao, Qi~Wu, Gordon Wetzstein, and Chelsea Finn.
\newblock {HumanPlus}: Humanoid shadowing and imitation from humans.
\newblock In \emph{Conference on Robot Learning (CoRL)}, 2024.
\newblock arXiv:2406.10454.

\bibitem[Han et~al.(2026)Han, Qiu, Bai, Xiao, Zeng, Liu, Yang, Jain, Ma, Fu,
  Zheng, Natarajan, Irshad, Shaw, Gombolay, Kira, and
  Ravichandar]{video2sim2real}
Yunhai Han, Jianuo Qiu, Linhao Bai, et~al.
\newblock {Video2Sim2Real}: Full-stack autonomous dexterous skill acquisition
  from a single human video.
\newblock \emph{arXiv preprint arXiv:2606.08828}, 2026.

\bibitem[He et~al.(2024)He, Luo, He, Xiao, Zhang, Zhang, Kitani, Liu, and
  Shi]{omnih2o}
Tairan He, Zhengyi Luo, Xialin He, et~al.
\newblock {OmniH2O}: Universal and dexterous human-to-humanoid whole-body
  teleoperation and learning.
\newblock In \emph{Conference on Robot Learning (CoRL)}, 2024.
\newblock arXiv:2406.08858.

\bibitem[He et~al.(2026)He, Xu, Li, Dong, Bian, Wang, and Gui]{ultra}
Xialin He, Sirui Xu, Xinyao Li, et~al.
\newblock {ULTRA}: Unified multimodal control for autonomous humanoid
  whole-body loco-manipulation.
\newblock \emph{arXiv preprint arXiv:2603.03279}, 2026.

\bibitem[Ji et~al.(2025)Ji, Peng, Liu, Li, Yang, Cheng, and Wang]{exbody2}
Mazeyu Ji, Xuanbin Peng, Fangchen Liu, et~al.
\newblock {ExBody2}: Advanced expressive humanoid whole-body control.
\newblock In \emph{Robotics: Science and Systems (RSS)}, 2025.
\newblock arXiv:2412.13196.

\bibitem[Jiang et~al.(2023)Jiang, Liu, Wang, and Wang]{grasptta}
Hanwen Jiang, Shaowei Liu, Jiashun Wang, and Xiaolong Wang.
\newblock {GraspTTA}: Rapidly adapting grippers for object grasping via
  test-time adaptation.
\newblock In \emph{Proc. Robotics: Science and Systems (RSS)}, 2023.

\bibitem[Jiang et~al.(2026)Jiang, Chen, Xu, Chen, Hao, and He]{egoalign}
Yiming Jiang, Jin Chen, Chongyang Xu, Yilun Chen, Aimin Hao, and Yisheng He.
\newblock {EgoAlign}: Bridging the human-humanoid gap for long-range
  loco-manipulation.
\newblock \emph{arXiv preprint arXiv:2609.38046}, 2026.

\bibitem[Kareer et~al.(2025)Kareer, Patel, Punamiya, Mathur, Cheng, Wang,
  Hoffman, and Xu]{egomimic}
Simar Kareer, Dhruv Patel, Ryan Punamiya, et~al.
\newblock {EgoMimic}: Scaling imitation learning via egocentric video.
\newblock In \emph{IEEE International Conference on Robotics and Automation
  (ICRA)}, 2025.
\newblock arXiv:2410.24221.

\bibitem[Kong et~al.(2025)Kong, Fort, Kang, Wittmer, et~al.]{ariagen2}
Chen Kong, James Fort, Aria Kang, Jonathan Wittmer, et~al.
\newblock {Aria Gen 2} pilot dataset.
\newblock \emph{arXiv preprint arXiv:2510.16134}, 2025.

\bibitem[{Lambda-humanoid Team} et~al.(2026){Lambda-humanoid Team}, Xu, Chen,
  Ye, Chen, Jiang, Wu, Wei, Qu, Zhang, He, Jiang, Meng, Wu, Xu, and
  Hoi]{lambda0}
{Lambda-humanoid Team}, Chongyang Xu, Jin Chen, et~al.
\newblock {$\lambda_0$}: Scaling open-world egocentric human data for humanoid
  loco-manipulation.
\newblock \emph{arXiv preprint arXiv:2610.18950}, 2026.

\bibitem[Lepert et~al.(2025{\natexlab{a}})Lepert, Fang, and Bohg]{masquerade}
Marion Lepert, Jiaying Fang, and Jeannette Bohg.
\newblock Masquerade: Learning from in-the-wild human videos using
  data-editing.
\newblock \emph{arXiv preprint arXiv:2508.09976}, 2025{\natexlab{a}}.

\bibitem[Lepert et~al.(2025{\natexlab{b}})Lepert, Fang, and Bohg]{phantom}
Marion Lepert, Jiaying Fang, and Jeannette Bohg.
\newblock Phantom: Training robots without robots using only human videos.
\newblock In \emph{Conference on Robot Learning (CoRL)}, 2025{\natexlab{b}}.
\newblock arXiv:2503.00779.

\bibitem[Li et~al.(2024)Li, Zhu, Xie, Jiang, Seo, Pavlakos, and Zhu]{okami}
Jinhan Li, Yifeng Zhu, Yuqi Xie, et~al.
\newblock {OKAMI}: Teaching humanoid robots manipulation skills through single
  video imitation.
\newblock In \emph{Conference on Robot Learning (CoRL)}, 2024.
\newblock arXiv:2410.11792.

\bibitem[Li et~al.(2026)Li, Yang, Miao, Zhu, Chen, Guan, Wu, Zhan, Sun, Huang,
  Shan, Jin, Hong, Wang, Feng, Liu, Wang, Yang, Zhou, Luo, Cheng, Zhang, Li,
  Cai, Yao, Zhang, Lin, Xu, Yu, Shen, Shen, and Li]{openaoe}
Zishuo Li, Bowen Yang, Changtao Miao, et~al.
\newblock {Open-AoE}: An open egocentric manipulation dataset and toolchain for
  embodied learning.
\newblock \emph{arXiv preprint arXiv:2607.14183}, 2026.

\bibitem[Lin et~al.(2024)Lin, Hu, Sheng, Wen, et~al.]{datascaling}
Fanqi Lin, Yingdong Hu, Pingyue Sheng, Chuan Wen, et~al.
\newblock Data scaling laws in imitation learning for robotic manipulation.
\newblock \emph{arXiv preprint arXiv:2410.18647}, 2024.

\bibitem[Lin et~al.(2026)Lin, Yang, Lian, Shen, Yu, Wu, Liu, Zhang, Li, Su,
  Liu, He, Shi, Huang, Zhang, Cheng, and Chen]{humanashumanoid}
Xiaopeng Lin, Ruoqi Yang, Shijie Lian, et~al.
\newblock {Human-as-Humanoid}: Enabling zero-shot humanoid learning from
  ego-exo human videos with human-aligned embodiments.
\newblock \emph{arXiv preprint arXiv:2606.32009}, 2026.

\bibitem[Liu et~al.(2025)Liu, Adeniji, Zhan, Haldar, et~al.]{egozero}
Vincent Liu, Ademi Adeniji, Haotian Zhan, Siddhant Haldar, et~al.
\newblock {EgoZero}: Robot learning from smart glasses.
\newblock \emph{arXiv preprint arXiv:2505.20290}, 2025.

\bibitem[Liu et~al.(2026)Liu, Cheng, Yin, Shin, Cueva, Yang, Chen, Zhang, and
  Xu]{egoengine}
Yangcen Liu, Shuo Cheng, Xinchen Yin, et~al.
\newblock {EgoEngine}: From egocentric human videos to high-fidelity dexterous
  robot demonstrations.
\newblock \emph{arXiv preprint arXiv:2606.12604}, 2026.

\bibitem[Lum et~al.(2025)Lum, Lee, Liu, and Bohg]{human2sim2robot}
Tyler Ga~Wei Lum, Olivia~Y. Lee, C.~Karen Liu, and Jeannette Bohg.
\newblock Crossing the human-robot embodiment gap with sim-to-real {RL} using
  one human demonstration.
\newblock \emph{arXiv preprint arXiv:2504.12609}, 2025.

\bibitem[Luo et~al.(2025)Luo, Feng, Zhang, Zheng, et~al.]{beingh0}
Hao Luo, Yicheng Feng, Wanpeng Zhang, Sipeng Zheng, et~al.
\newblock {Being-H0}: Vision-language-action pretraining from large-scale human
  videos.
\newblock \emph{arXiv preprint arXiv:2507.15597}, 2025.

\bibitem[Luo et~al.(2026)Luo, Yuan, Wang, Li, Casta{\~n}eda, Chen, Cao, Li,
  Minor, Ben, Park, Sami, Wang, Da, Ding, Hogg, Song, Lim, Jeong, He, Xue,
  Xiao, Yuen, Kautz, Chang, Iqbal, Fan, and Zhu]{sonic}
Zhengyi Luo, Ye~Yuan, Tingwu Wang, et~al.
\newblock {SONIC}: Supersizing motion tracking for natural humanoid whole-body
  control.
\newblock \emph{Science Robotics}, 11\penalty0 (117):\penalty0 eaed4592, 2026.
\newblock arXiv:2511.07820.

\bibitem[Nai et~al.(2026)Nai, Zheng, Zhao, Zhu, Dai, Chen, Hu, Hu, Zhang, Wen,
  and Gao]{humi}
Ruiqian Nai, Boyuan Zheng, Junming Zhao, et~al.
\newblock {HuMI}: Humanoid whole-body manipulation from humanoid-free
  demonstrations.
\newblock \emph{arXiv preprint arXiv:2602.06643}, 2026.

\bibitem[{NVIDIA}(2026)]{groot17}
{NVIDIA}.
\newblock {NVIDIA Isaac GR00T N1.7 (3B)}.
\newblock Model card, Hugging Face model ID \texttt{nvidia/GR00T-N1.7-3B},
  2026.

\bibitem[{NVIDIA} et~al.(2025){NVIDIA}, Bjorck, Casta{\~n}eda, Cherniadev, Da,
  Ding, Fan, Fang, Fox, Hu, Huang, Jang, Jiang, Kautz, Kundalia, Lao, Li, Lin,
  Lin, Liu, Llontop, Magne, Mandlekar, Narayan, Nasiriany, Reed, Tan, Wang,
  Wang, Wang, Wang, Xiang, Xie, Xu, Xu, Ye, Yu, Zhang, Zhang, Zhao, Zheng, and
  Zhu]{groot}
{NVIDIA}, Johan Bjorck, Fernando Casta{\~n}eda, et~al.
\newblock {GR00T N1}: An open foundation model for generalist humanoid robots.
\newblock \emph{arXiv preprint arXiv:2503.14734}, 2025.

\bibitem[{Physical Intelligence} et~al.(2025){Physical Intelligence}, Black,
  Brown, Darpinian, Dhabalia, Driess, Esmail, Equi, Finn, Fusai, Galliker,
  Ghosh, Groom, Hausman, Ichter, Jakubczak, Jones, Ke, LeBlanc, Levine,
  Li-Bell, Mothukuri, Nair, Pertsch, Ren, Shi, Smith, Springenberg, Stachowicz,
  Tanner, Vuong, Walke, Walling, Wang, Yu, and Zhilinsky]{pi05}
{Physical Intelligence}, Kevin Black, Noah Brown, et~al.
\newblock $\pi_{0.5}$: A vision-language-action model with open-world
  generalization.
\newblock \emph{arXiv preprint arXiv:2504.16054}, 2025.

\bibitem[Qiu et~al.(2025)Qiu, Yang, Cheng, Chawla, et~al.]{hat}
Ri-Zhao Qiu, Shiqi Yang, Xuxin Cheng, Chaitanya Chawla, et~al.
\newblock Humanoid policy $\sim$ human policy.
\newblock \emph{arXiv preprint arXiv:2503.13441}, 2025.

\bibitem[Rempe et~al.(2026)Rempe, Petrovich, Yuan, Zhang, Peng, Jiang, Wang,
  Iqbal, Minor, de~Ruyter, Li, Tessler, Lim, Jeong, Wu, Hassani, Huang, Yu,
  Chung, Song, Dionne, Kautz, Yuen, and Fidler]{kimodo}
Davis Rempe, Mathis Petrovich, Ye~Yuan, et~al.
\newblock {Kimodo}: Scaling controllable human motion generation.
\newblock \emph{arXiv preprint arXiv:2603.15546}, 2026.

\bibitem[Shi et~al.(2026)Shi, Peng, Chen, Jiang, Li, Huang, Luo, Li, and
  Chen]{egohumanoid}
Modi Shi, Shijia Peng, Jin Chen, et~al.
\newblock {EgoHumanoid}: Unlocking in-the-wild loco-manipulation with
  robot-free egocentric demonstration.
\newblock In \emph{Robotics: Science and Systems (RSS)}, 2026.
\newblock arXiv:2602.10106.

\bibitem[Wang et~al.(2026{\natexlab{a}})Wang, Ren, Morgan, Chen, Qian,
  Chanrungmaneekul, and Hang]{egoinfinity}
Gaotian Wang, Kejia Ren, Andrew Morgan, et~al.
\newblock {EgoInfinity}: A web-scale {4D} hand-object interaction data engine
  for any-view robot retargeting and video-to-action robot learning.
\newblock \emph{arXiv preprint arXiv:2606.17385}, 2026{\natexlab{a}}.

\bibitem[Wang et~al.(2023)Wang, Zhang, Chen, Xu, Li, Liu, and
  Wang]{dexgraspnet}
Ruicheng Wang, Jialiang Zhang, Jiayi Chen, et~al.
\newblock {DexGraspNet}: A large-scale robotic dexterous grasp dataset for
  general objects based on simulation.
\newblock In \emph{Proc. IEEE Int. Conf. Robot. Autom. (ICRA)}, 2023.

\bibitem[Wang et~al.(2026{\natexlab{b}})Wang, Lin, Chen, Yuan, Liang, Huang,
  Chen, Lei, Zhang, Zhang, Li, Zhang, Xiao, Jiao, and Jin]{ego2robot}
Ye~Wang, Pei Lin, Xiong-Hui Chen, et~al.
\newblock {Ego2Robot}: Scalable robot data synthesis from egocentric human
  data.
\newblock \emph{arXiv preprint arXiv:2608.02580}, 2026{\natexlab{b}}.

\bibitem[Weng et~al.(2025)Weng, Li, Sobanbabu, Wang, Luo, He, Ramanan, and
  Shi]{hdmi}
Haoyang Weng, Yitang Li, Nikhil Sobanbabu, et~al.
\newblock {HDMI}: Learning interactive humanoid whole-body control from human
  videos.
\newblock \emph{arXiv preprint arXiv:2509.16757}, 2025.

\bibitem[Xie et~al.(2026)Xie, Zhang, Park, Wang, Wen, Li, Li, Ben, Weng, Ye,
  Minor, Wang, Jiang, Fidler, Kautz, Fan, Zhu, Luo, Iqbal, and Yuan]{grail}
Tianyi Xie, Haotian Zhang, Jinhyung Park, et~al.
\newblock {GRAIL}: Generating humanoid loco-manipulation from world models.
\newblock \emph{arXiv preprint arXiv:2606.05160}, 2026.

\bibitem[Xue et~al.(2025)Xue, Huang, Niu, Liao, et~al.]{leverb}
Haoru Xue, Xiaoyu Huang, Dantong Niu, Qiayuan Liao, et~al.
\newblock {LeVERB}: Humanoid whole-body control with latent vision-language
  instruction.
\newblock \emph{arXiv preprint arXiv:2506.13751}, 2025.

\bibitem[Yang et~al.(2026)Yang, Huang, Wu, Kanazawa, Abbeel, Sferrazza, Liu,
  Duan, and Shi]{omniretarget}
Lujie Yang, Xiaoyu Huang, Zhen Wu, et~al.
\newblock {OmniRetarget}: Interaction-preserving data generation for humanoid
  whole-body loco-manipulation and scene interaction.
\newblock In \emph{IEEE International Conference on Robotics and Automation
  (ICRA)}, 2026.
\newblock arXiv:2509.26633.

\bibitem[Yang et~al.(2025)Yang, Yu, Wu, Yan, et~al.]{egovla}
Ruihan Yang, Qinxi Yu, Yecheng Wu, Rui Yan, et~al.
\newblock {EgoVLA}: Learning vision-language-action models from egocentric
  human videos.
\newblock \emph{arXiv preprint arXiv:2507.12440}, 2025.

\bibitem[Ze et~al.(2025)Ze, Chen, Araujo, Cao, Peng, Wu, and Liu]{twist}
Yanjie Ze, Zixuan Chen, Joao~Pedro Araujo, et~al.
\newblock {TWIST}: Teleoperated whole-body imitation system.
\newblock \emph{arXiv preprint arXiv:2505.02833}, 2025.

\bibitem[Zhang et~al.(2026)Zhang, Ma, Liu, Shi, Su, Wang, Sun, Cui, Yu, Han,
  Zhao, Sun, Yin, Wang, Cao, Zhang, Cheng, Hao, Ji, Liang, Tang, Xu, and
  Guo]{meshmimic}
Qiang Zhang, Jiahao Ma, Peiran Liu, et~al.
\newblock {MeshMimic}: Geometry-aware humanoid motion learning through {3D}
  scene reconstruction.
\newblock \emph{arXiv preprint arXiv:2602.15733}, 2026.

\bibitem[Zhou et~al.(2026)Zhou, Chen, Wu, Xie, Xu, Yin, Wang, Shi, Chen, Bai,
  and Li]{koda}
Zibo Zhou, Yueru Chen, Junsong Wu, et~al.
\newblock {KODA}: Kinematics--observation data alignment for humanoid learning
  from human egocentric demonstration.
\newblock \emph{arXiv preprint arXiv:2610.21152}, 2026.

\end{thebibliography}

\clearpage
\appendices
\crefalias{section}{appendix}

\section{Agentic Real-to-Sim Details}
\label{app:r2s}

\paragraph{Asset frames and support fitting}
Each reconstructed asset $o$ has a metric local frame with world pose
$T^{\text{obj}}_o(t)$ (constant for static supports and background) and carries
a visual mesh $\mathcal{M}^{\text{V}}_o$ and collision model
$\mathcal{M}^{\text{C}}_o$ attached by fixed local transforms $T^{o}_{\text{V}}$
and $T^{o}_{\text{C}}$,
\begin{equation}
  T^{\text{V}}_o(t)=T^{\text{obj}}_o(t)\,T^{o}_{\text{V}},\qquad
  T^{\text{C}}_o(t)=T^{\text{obj}}_o(t)\,T^{o}_{\text{C}},
  \label{eq:asset}
\end{equation}
so appearance and contact geometry cannot drift apart. Before fusing geometry,
the agent verifies whether the published stereo depth maps $D_k$ encode axial
depth along the optical axis or radial ray range by comparing cross-view
reprojection residuals, tracked feature agreement and bare-floor planarity. A
robust plane fit to the segmented metric points $\mathcal{Q}_s$ of each support
region $s$,
\begin{equation}
  \Pi_s=\arg\min_{\Pi}\ \sum_{\mathbf{x}\in\mathcal{Q}_s}\varrho\big(\operatorname{dist}(\mathbf{x},\Pi)\big),
  \label{eq:support}
\end{equation}
with robust penalty $\varrho$, fixes the support height and tilt, while its
observed boundary edges fix the in-plane origin, yaw and footprint. The floor is
fitted from visible bare-floor regions in the same way and defines $z=0$ of the
gravity-aligned world frame.

\paragraph{Object-frame TSDF fusion and texture baking}
Geometry and appearance are reconstructed by fusing valid depth observations
into a truncated signed distance field~\cite{curless1996} expressed in each
asset's local frame $T^{\text{obj}}_o(c_k)$,
\begin{align}
  F_o(\mathbf{v})&=\frac{\sum_k\omega_{o,k}(\mathbf{v})\,\phi_k\big(T^{\text{obj}}_o(c_k)\,\mathbf{v}\big)}{\sum_k\omega_{o,k}(\mathbf{v})},\nonumber\\
  \phi_k(\mathbf{x})&=\operatorname{clip}\Big(\frac{D_k\big(\pi_k(\mathbf{x})\big)-z_k(\mathbf{x})}{\mu},\,-1,\,1\Big),
  \label{eq:tsdf}
\end{align}
where $\pi_k$ projects a world point $\mathbf{x}$ into view $k$, $z_k(\mathbf{x})$
is its axial depth, $\mu$ is the truncation distance, and $\omega_{o,k}(\mathbf{v})=1$
only where depth is valid, the pixel belongs to asset $o$ and the pose at $c_k$
is reliable. Multi-view masks reviewed in native fisheye coordinates are lifted
through calibrated depth and visibility gates to partition the fused mesh into
disjoint face sets for each object and its complementary support surface.

Each face $f$ of the visual mesh $\mathcal{M}^{\text{V}}_o$ takes its colour
from the source view $k^{*}(f)$ that maximizes visibility, normal alignment and
depth consistency,
\begin{equation}
  k^{*}(f)=\arg\max_k\ \nu_k(f)\,m_k(f)\,
  \frac{\cos^{2}\vartheta_{fk}}{d_{fk}^{2}+\epsilon}\,
  \exp\Big(-\frac{r_{fk}^{2}}{\sigma_r^{2}}\Big),
  \label{eq:texture}
\end{equation}
where $\nu_k(f)=1$ when face $f$ lies inside the camera frustum with depth
residual $r_{fk}$ consistent with $D_k$, $m_k(f)=0$ when occluded by a hand,
the body or another object, $\vartheta_{fk}$ is the viewing angle relative to
the face normal, and $d_{fk}$ is the camera distance. Faces without a valid
source view retain their fused vertex colour.

\paragraph{Bounded geometric and appearance completion}
When an extracted object is only partly observed or moving it exposes an
unobserved region of its support surface, the agent constructs a separately
labeled completion candidate while preserving the observed-only asset. For a
compact object, a bounded low-parameter surface (such as a superellipsoid) is
fitted by robust least squares to the observed metric vertices and multi-view
silhouette masks inside the object's local frame while keeping the external pose
transform $T^{\text{obj}}_o$ unchanged. Source RGB textures are projected onto
front-facing triangles whose vertices and centres fall inside the object mask,
and unseen faces receive median material colour. For partially scanned supports
or furniture, the observed textured mesh is retained and paired with a
separately addressable inset backing shell extracted from the signed-distance
union of bounded structural volumes (base, seat support, back and arms). Added
backing faces receive spatially and normal-weighted colour interpolated from
neighbouring observed vertices, with distant unseen regions falling back to
median upholstery colour. Completed visual shells are flagged as inferred in
asset provenance and kept distinct from the validated collision model
$\mathcal{M}^{\text{C}}_o$.

\paragraph{Contact events and target contact bands}
From the tracked hand and object trajectories (without relying on body
tracking), the agent detects interaction events, namely reach onset (when the
arm begins extending), first touch (when a fingertip approaches within
\SI{1}{\centi\meter} of its closest distance to the object), full grasp
closure, lift-off (when the object moves more than \SI{1}{\centi\meter} from
its touch pose) and release. Rather than labeling isolated fingertip points,
the agent localizes a coherent target contact band
$\mathcal{R}_m\subset\partial\mathcal{M}^{\text{C}}_o$ in the object's local
coordinate frame around the grasp-closure anchor frame. Candidate surface
points on the object mesh are projected into calibrated views to score
proximity to observed hand masks (or evaluated against synchronized 3D hand
samples transformed into the object frame via
$\mathbf{p}_{\text{obj}}=R^{\top}(\mathbf{p}_{\text{world}}-\mathbf{t})$). The
reviewed seed location is expanded into a surface band (such as an angular
sector across inner/outer walls and rim, an arc along a handle or rail, or a
geodesically expanded patch on an irregular mesh) with subdivided triangles and
inner/outer face masks stored in the object frame.

\section{Motion Adaptation Details}
\label{app:adapt}

\paragraph{Head residual, root reference and rule-based grasp planning}
In \cref{eq:problem}, the head-tracking residual keeps the components that a
camera fixed to the robot torso can reproduce,
\begin{equation}
  e^{\text{head}}(x,h)=
  \begin{bmatrix}
  p^{\text{cam}}_{xy}(x)-p^{\text{head}}_{xy}(h)\\
  \psi^{\text{cam}}(x)-\psi^{\text{head}}(h)\\
  \Delta z^{\text{cam}}(x)-\rho\,\Delta z^{\text{head}}(h)
  \end{bmatrix},
  \label{eq:head}
\end{equation}
where $\rho$ is the ratio of robot to human standing camera height. Given the
demonstrator's head position $p^{\text{head}}(h)$ and horizontal viewing
direction $f(h)$, the planar root reference $(p^{*}(h),\psi^{*}(h))$ projects
the head trajectory onto the floor plane and smooths the heading,
\begin{align}
  p^{*}(h)&=p^{\text{head}}_{xy}(h),\nonumber\\
  \psi^{*}(h)&=\big(g_\sigma\ast\operatorname{unwrap}\operatorname{atan2}(f_y,f_x)\big)(h),
  \label{eq:root}
\end{align}
where $g_\sigma$ is a Gaussian kernel that suppresses brief head turns. During
contact $m$ with object $o=o_m$ (\cref{eq:anchor}), a rule-based grasp planner
solves for the task-arm and 7-DoF Dex3 finger joints (yielding the object-frame
wrist pose $T^{\text{grasp}}_a$ and closed finger posture $q^{\text{closed}}_a$)
by bounded least-squares optimization over MuJoCo geom-to-geom signed
distances,
\begin{equation}
  \begin{aligned}
  \min_{\theta^{\text{arm}},\,q^{\text{dex}}}\ &\sum_{r}w_r\,\psi_r\big(\theta^{\text{arm}},q^{\text{dex}};\,\mathcal{R}_m,\mathcal{M}^{\text{C}}_o\big)\\
  &+\big\|T^{\text{wrist}}(\theta^{\text{arm}})\ominus T^{\text{hum}}_a\big\|_{W_g}^{2},
  \end{aligned}
  \label{eq:grasp}
\end{equation}
where each residual $\psi_r$ is a hinge penalty (zero inside its tolerance and
quadratic outside) or soft preference defined in the local frame of the contact
band $\mathcal{R}_m$ (with outward normal $\mathbf{n}$, tangent $\mathbf{t}$ and
vertical axis $\mathbf{z}$). The rules align the hand width along $\mathbf{t}$
with the palm facing the band, seat the rim or edge inside the thumb--index web
without deep penetration, place the index and middle fingertip pads onto the
outer face of $\mathcal{R}_m$ and the thumb onto the opposing inner or near
face, keep all other hand and arm links clear of the object, support and robot
body, centre the palm on $\mathcal{R}_m$, and keep palm and forearm orientations
within ${\sim}12^{\circ}\text{--}15^{\circ}$ of the human demonstration at full
grasp closure. For simple cylindrical objects resting on a support, the target
slides the observed palm pose along the object axis until the open Dex3 hand
clears the support surface. The solver runs from multiple initializations and
retains the lowest-cost solution whose fingertip pads lie within \SI{4}{\milli\meter}
of $\mathcal{R}_m$ with no link penetrating beyond \SI{3}{\milli\meter}. In the
bilevel stance optimization of \cref{eq:keyframe,eq:stance}, the outer problem
over $s=(p_s,\psi_s)\in\mathrm{SE}(2)$ is solved by grid search followed by
local gradient refinement around the projected root position, while the inner
problem \cref{eq:keyframe} is solved by damped least-squares inverse kinematics.

\paragraph{Retiming and windowed generation}
A timeline segment of human duration $\Delta h_n=t_{n+1}-t_n$ receives robot
duration $D_n$ and piecewise-linear time map $h(\tau)$,
\begin{align}
  D_n&=\max\Big(\Delta h_n,\ \max_{j}\ell_{n,j}/v_j\Big),\label{eq:duration}\\
  h(\tau)&=t_n+\frac{\tau-\tau_n}{D_n}\,\Delta h_n,\quad
  \tau\in[\tau_n,\tau_n+D_n),
  \label{eq:time-map}
\end{align}
where $\ell_{n,j}$ is the root or end-effector path length over segment $n$,
$v_j$ is the robot speed budget, and $\tau_{n+1}=\tau_n+D_n$. The
constraint-conditioned motion generator $G$~\cite{kimodo} synthesizes
whole-body motion in windows $w$ of length $N_w$,
\begin{equation}
  \hat x^{w}_{1:N_w}=G\big(\ell_w;\,\mathcal{C}_w\big),
  \label{eq:prior}
\end{equation}
conditioned on root waypoints, tracked hand keyframes, stance keyframes and an
$H$-frame overlap constraint $\hat x^{w+1}_l=\hat x^w_{N_w-H+l}$ ($l=1,\dots,H$).
Adjacent windows are blended across the overlap and the boundary residual
$\Delta=\bar x_H\ominus\hat x^{w+1}_H$ is decayed over the subsequent $L$ frames,
\begin{align}
  \bar x_l&=\operatorname{interp}\Big(\hat x^{w}_{N_w-H+l},\,\hat x^{w+1}_{l};\,\tfrac{l}{H+1}\Big),\nonumber\\
  \hat x^{w+1}_{H+l}&\leftarrow \hat x^{w+1}_{H+l}\oplus\Big(1-\tfrac{l}{L+1}\Big)\Delta.
  \label{eq:splice}
\end{align}

\section{Motion Correction and Screening Details}
\label{app:corr}

\paragraph{Fade ramp and kinematic refinement}
Across the fade-in window $[t^{\text{in}}_{a,k}-T_{\text{f}},\,t^{\text{in}}_{a,k}]$
and fade-out window $[t^{\text{out}}_{a,k},\,t^{\text{out}}_{a,k}+T_{\text{f}}]$
of tracked segment $k$, the blended hand target $T^{*}_a(h)$ is given by
\begin{align}
  T^{*}_a(h)&=\hat T_a(h)\oplus b_{a,k}(h)\,\big[T^{\text{obs}}_a(\bar h)\ominus\hat T_a(\bar h)\big],\nonumber\\
  b_{a,k}(h)&=\beta\Big(1+\frac{h-t^{\text{in}}_{a,k}}{T_{\text{f}}}\Big)\,
  \beta\Big(1+\frac{t^{\text{out}}_{a,k}-h}{T_{\text{f}}}\Big),
  \label{eq:fade}
\end{align}
with $\bar h=\operatorname{clip}(h,t^{\text{in}}_{a,k},t^{\text{out}}_{a,k})$,
$\beta(u)=\operatorname{clip}(3u^2-2u^3,0,1)$ and $T_{\text{f}}=0.6$\,s; the fade-in thus ends exactly when the hand enters view. After resampling
$\hat x_i=(\hat p_i,\hat R_i,\hat\theta_i)$ to the control rate, kinematic
refinement solves
\begin{equation}
  \min_{u_{1:N}\in\mathcal{U}}\ \sum_{i}\Big(J^{\text{task}}_i+J^{\text{prior}}_i
  +J^{\text{stance}}_i+J^{\text{bal}}_i+J^{\text{coll}}_i+J^{\text{smooth}}_i\Big),
  \label{eq:kin}
\end{equation}
over $u_i=(\delta p_i,\delta\omega_i,\theta_i)$ with $p_i=\hat p_i+\delta p_i$
and $R_i=\hat R_i\operatorname{Exp}(\delta\omega_i)$, where
\begin{align*}
  J^{\text{task}}_i&=\textstyle\sum_{a\in\mathcal{A}(i)}\big\|T_a(u_i)\ominus T^{*}_a(h_i)\big\|^{2}_{W_{\text{task}}},\\
  J^{\text{prior}}_i&=\|\theta_i-\hat\theta_i\|^{2}_{W_\theta(i)}+\|\delta p_i\|^{2}_{W_p}+\|\delta\omega_i\|^{2}_{W_\omega},\\
  J^{\text{stance}}_i&=\textstyle\sum_{f\in\mathcal{F}(i)}\big\|T_f(u_i)\ominus\bar T_f\big\|^{2}_{W_f},\\
  J^{\text{bal}}_i&=\big[\varepsilon_b-d\big(\operatorname{proj}_{xy}(r^{\text{com}}_i),\mathcal{P}_i\big)\big]_{+}^{2}/\sigma_b^{2},\\
  J^{\text{coll}}_i&=\textstyle\sum_{(g,j)}\big[\varepsilon_c-d_{gj}(u_i,h_i)\big]_{+}^{2}/\sigma_c^{2},\\
  J^{\text{smooth}}_i&=\big\|(u_{i+1}-\hat u_{i+1})-(u_i-\hat u_i)\big\|^{2}_{W_s}.
\end{align*}
The prior standard deviation for joint group $g$ interpolates log-linearly as
$\sigma_g(h)=(\sigma^{\text{ref}}_g)^{1-\alpha_g(h)}(\sigma^{\text{task}}_g)^{\alpha_g(h)}$
with $\alpha_g(h)=\max_{(a,k):\,g\in\operatorname{grp}(a)}b_{a,k}(h)$ and
$W_\theta(i)=\operatorname{diag}(\sigma_g(h_i)^{-2})$.

\paragraph{Hand closure and rollout screening}
The 7-DoF Dex3 hand command is computed as
\begin{align}
  q^{\text{dex}}(t)&=q^{\text{open}}+c(t)\big(q^{\text{closed}}-q^{\text{open}}\big),\nonumber\\
  c(t)&=c^{\star}_{m}\,\big(\rho_c\ast\lambda\big)(t),\quad
  \lambda(t)=\mathbf{1}\big[t\in W_m\big]\,\mathbf{1}\big[\mu(t)\ge\mu_0\big],
  \label{eq:hand}
\end{align}
where $W_m=[t^{\text{on}}_m,t^{\text{off}}_m]$, $\mu(t)$ is the pinch and curl
statistic from landmarks $L_a$, and $\rho_c$ is a smoothing kernel. A physics
rollout passes screening when $\Lambda=1$,
\begin{align}
  \Lambda&=\mathbf{1}\big[\text{progress}\big]\,
  \mathbf{1}\big[e^{\text{obj}}_N\le\epsilon_{\text{obj}}\big]\,
  \mathbf{1}\big[\operatorname{med}_i e^{\text{head}}_i\le\epsilon_{\text{head}}\big]\nonumber\\
  &\quad\cdot\prod_{j}\mathbf{1}\big[\zeta_j\le\epsilon_j\big],
  \label{eq:screen}
\end{align}
where $\zeta_j$ checks torso tilt, peak joint velocity, high-frequency joint
acceleration, stance-foot slip and pelvis height drop.

\section{Rewards and Rendering Formulation}
\label{app:reward}
\label{app:render}

With the Gaussian kernel $\kappa(e;\varsigma)=\exp(-\|e\|^{2}/\varsigma^{2})$,
a coding agent composes the step reward from a term library $\mathcal{K}(\mathcal{E})$
based on the per-task semantics, detected contact events $\mathcal{E}$ and
predicted contact bands $\mathcal{R}_m$,
\begin{align}
  r_i&=\sum_{k\in\mathcal{K}(\mathcal{E})}w_k\,r_{k,i},\label{eq:reward}\\
  r^{\text{ev}}_{m,i}&=\mathbf{1}\big[i\in\mathcal{W}_m\big]\,
  \kappa\big(\max(d_{m,i}-d_0,0);\varsigma_d\big)\,
  \kappa\big(\vartheta_{m,i};\varsigma_\vartheta\big),\label{eq:event}
\end{align}
where $d_{m,i}$ and $\vartheta_{m,i}$ measure the object's position and axis
alignment errors relative to its target pose in the world frame or in a target
body link frame (such as the head frame when seating a hat outside the camera's
field of view) during manipulation window $\mathcal{W}_m$.

Along an accepted rollout, each pixel $u$ of the deployed robot camera model
$\pi_{\text{rob}}$ casts ray $r_i(u)=T^{\text{cam}}(s_i)\,\pi_{\text{rob}}^{-1}(u)$
into the composite scene
$\mathcal{M}^{\text{V}}_{\text{bg}}\cup\chi_i(\mathcal{M}^{\text{V}}_o)\cup\mathcal{L}(s_i)$.
Writing $\xi_i(u)$ for the first surface intersection and $C$ for the baked
surface or robot link material colour, the rendered pixel is
$I_i(u)=C(\xi_i(u))$ when $\xi_i(u)$ exists and $\varnothing$ otherwise.

\end{document}